\documentclass[journal, comsoc]{IEEEtran}
\usepackage[T1]{fontenc}

\usepackage{xcolor}
\usepackage{dsfont}
\usepackage{amsbsy}
\usepackage{amssymb}
\usepackage{graphicx}
\usepackage{bbm}
\usepackage{subcaption}
\ifCLASSINFOpdf

\else

\fi

\usepackage{multirow}
\usepackage{float}
\usepackage{hyperref}
\usepackage[export]{adjustbox}
\usepackage{xcolor}
\usepackage{caption}
\usepackage{subcaption}
\usepackage{afterpage}
\usepackage[labelsep=period, figurename=Figure]{caption}

\usepackage{amsmath}
\usepackage{verbatim}
\usepackage{graphicx}

\usepackage{tikz}
\usepackage{textcomp}
\usepackage{hyperref}
\usepackage{lipsum}
\usepackage{comment}
\usepackage{algpseudocode}
\usepackage{amsmath}
\usepackage{algorithm}
\usepackage{cite}
\usepackage{graphicx}
\usepackage{textcomp}
\usepackage{xcolor}
\usepackage{amsmath,amssymb,amsfonts}
\usepackage{hyperref}
\usepackage{booktabs}

\usepackage{enumitem}
\setlist[enumerate,1]{label={\arabic*.}}
\makeatother

\IEEEoverridecommandlockouts

\begin{document}
\addtolength{\textfloatsep}{-2.19pt}




\title{Trustworthy and Privacy-Aware Task-Oriented Semantic Communication for 6G-IoT Using Contrastive and Disentangled Representations}

\title{Trustworthy Semantic Communication via Contrastive Learning and Disentangled Representations for Privacy-Aware 6G-IoTs}

\title{Contrastive Disentanglement for Secure 6G Semantic JSCC with Built-in Privacy}

\title{Contrastive Learning and Adversarial Disentanglement for Privacy-Aware Task-Oriented Semantic Communication}

		        

\author{
         Omar~Erak,~\IEEEmembership{Member,~IEEE,} Omar~Alhussein,~\IEEEmembership{Senior~Member,~IEEE,}
         Wen~Tong,~\IEEEmembership{Fellow,~IEEE}
				\thanks{Omar Erak and Omar Alhussein are with the KU 6G Research Center, Department of Computer Science, Khalifa University, Abu Dhabi, UAE (e-mail: omarerak@ieee.org, omar.alhussein@ku.ac.ae).}
                \thanks{Wen Tong is with the Huawei Wireless Research, Wireless Advanced System
and Competency Centre, Huawei Technologies Co. Ltd., Ottawa,
ON K2K 3J1, Canada, (e-mail:
tongwen@huawei.com).}

		\vspace{-0.7cm}		}

\maketitle



\begin{abstract}
Task-oriented semantic communication systems have emerged as a promising approach to achieving efficient and intelligent data transmission in next-generation networks, where only information relevant to a specific task is communicated. This is particularly important in 6G-enabled Internet of Things (6G-IoT) scenarios, where bandwidth constraints, latency requirements, and data privacy are critical. However, existing methods struggle to fully disentangle task-relevant and task-irrelevant information, leading to privacy concerns and suboptimal performance. To address this, we propose an information-bottleneck inspired method, named CLAD (\underline{c}ontrastive \underline{l}earning and \underline{a}dversarial \underline{d}isentanglement). CLAD utilizes contrastive learning to effectively capture task-relevant features while employing adversarial disentanglement to discard task-irrelevant information. Additionally, due to the absence of reliable and reproducible methods to quantify the minimality of encoded feature vectors, we introduce the Information Retention Index (IRI), a comparative metric used as a proxy for the mutual information between the encoded features and the input. The IRI reflects how minimal and informative the representation is, making it highly relevant for privacy-preserving and bandwidth-efficient 6G-IoT systems. Extensive experiments demonstrate that CLAD outperforms state-of-the-art baselines in terms of semantic extraction, task performance, privacy preservation, and IRI, making it a promising building block for responsible, efficient and trustworthy 6G-IoT services.
\end{abstract}


\IEEEpeerreviewmaketitle

\section{Introduction}

In conventional communication systems, the primary objective has been to ensure reliable transmission of data, focusing on delivering bit sequences across noisy channels without considering the meaning, context, or purpose of the data being transmitted. Shannon's mathematical theory of communication focuses on optimizing metrics such as data rate, error rate, and bandwidth efficiency, whilst being agnostic to the ultimate purpose and relevance of the transmitted information \cite{shannon}. This approach has been widely successful and effective for general communication needs thus far. However, next-generation communication systems, beginning with 6G, require more intelligent and task-aware communication methods to support a wide range of real-time and mission-critical Internet of Things (IoT) applications \cite{6GVision, giordani2020toward}, such as computer vision \cite{voulodimos2018deep}, autonomous driving \cite{yaqoob2019autonomous}, extended reality (XR) \cite{andrade2019extended}, and generative artificial intelligence (AI) \cite{10384630}.

As we move towards these advanced systems, there is a growing recognition that communication should not merely be about transmitting raw data, but about understanding the underlying meaning and purpose of the data. This shift towards task-oriented semantic communication represents a fundamental change in the design of communication networks \cite{9991044, shi2023task}. Instead of focusing solely on the accurate and efficient transmission of bits, these new approaches aim to ensure that the information most relevant to the specific task or decision-making process is prioritized and delivered with minimal delay and overhead. This is especially critical in 6G-IoT settings, where devices operate under tight bandwidth, latency, and energy constraints, and where privacy and reliability are essential. For example, in a smart city environment \cite{mehmood2017internet}, rather than transmitting all sensor data from traffic cameras, task-oriented communication focuses on sending only the information necessary to identify and respond to potential hazards or optimize traffic flow in real time.

With the growing success and popularity of deep learning (DL) in various wireless communication applications \cite{mao2018deep, erak2023accelerating}, many emerging task-oriented communication systems have adopted DL approaches to encode task-relevant information to improve task performance and efficiency of the communication system \cite{shao2021learning, bourtsoulatze2019deep, xie2023robust}. Nevertheless, most proposed schemes do not focus on quantifying or benchmarking the amount of information that the encoded features retain about the input, primarily due to the computational difficulty of estimating mutual information and the lack of a unified methodology that provides fair and reproducible results. This omission is critical, particularly in IoT and edge computing scenarios, where understanding how much information is kept and whether it is necessary or private has direct implications on trustworthiness, data security, and system interpretability. 

Furthermore, most current approaches rely on maximizing mutual information between the encoded features and the target using variational approximations based on the cross-entropy loss \cite{alemi2017deep, shao2021learning, xie2023robust}. However, deriving a maximization for the mutual information between the encoded feature vector and the targets based on contrastive learning \cite{oord2018representation} remains unexplored for task-oriented communication systems.

To address the aforementioned challenges, we develop a task-oriented communication system based on contrastive learning \cite{oord2018representation} and disentangled representation learning \cite{paige2017learning}, and we devise a new metric to compute comparative values for a proxy of mutual information between the encoded features and the inputs across different systems, rather than computing the exact mutual information. More specifically, our major contributions are as follows:

\begin{itemize}
    \item We derive a lower bound for the mutual information between the encoded features and the target using contrastive learning principles. We show that the contrastive learning based lower bound improves task accuracy and performance compared to traditional cross-entropy based mutual information approximations;

    \item We propose a systematic training methodology based on an innovative loss function, designed to extract task-irrelevant information through reconstruction losses while disentangling it from task-relevant features using adversarial methods. This enables the system to prioritize the transmission of task-relevant features while minimizing communication overhead and reducing unnecessary information transmission, thus enhancing privacy;

    \item To address the current limitation of lacking a reliable and unified approach for estimating the mutual information between the encoded features and input data, we introduce a new metric named the Information Retention Index (IRI), which serves as a proxy for the mutual information. This metric compares the informativeness and minimality of the encoded features across various task-oriented communication methods, providing deeper insights into system behavior and enabling a more rigorous comparison of their performance;

    \item We evaluate our proposed task-oriented communication system in diverse channel conditions. It is tested against several existing task-oriented communication methods, demonstrating improved task performance, enhanced privacy awareness, and reduced amount of irrelevant information across a wide range of transmission scenarios.
\end{itemize}

\section{Related Work}

 DL-based communication systems have shown success in recent years. DeepJSCC (Deep Joint Source-Channel Coding) is a recent advancement in the field of wireless communication that utilizes deep learning to jointly optimize source and channel coding, which are traditionally treated as separate tasks \cite{bourtsoulatze2019deep}. Unlike conventional methods that rely on separate compression and error-correction codes, DeepJSCC uses neural networks to directly map source data to channel symbols, allowing for an end-to-end optimization of the communication system. DeepJSCC can be trained on a classification task by minimizing the cross-entropy loss, ensuring task-specific performance; however, it does not inherently ensure that only task-relevant features are transmitted.

Building on that, Shao et al. \cite{shao2021learning} proposed a task-oriented communication system for edge inference by leveraging the information bottleneck (IB) theory \cite{tishby2000information}, and variational approximations \cite{alemi2017deep} to balance a trade-off between the minimality of the transmitted feature vector and the task performance. Their results demonstrate improved latency and classification accuracy. Another work focuses on improving the aforementioned IB framework for task-oriented communication systems by introducing an information bottleneck framework that ensures robustness to varying channel conditions~\cite{xie2023robust}. 

Wang et al. \cite{wang2024privacy} formulate a privacy-utility trade-off to develop an IB-based privacy-preserving task-oriented communication system against model inversion attacks \cite{fredrikson2015model}. This is achieved by striking a balance between the traditional IB-based loss functions similar to the work discussed above and a mean squared error (MSE) based term that aims at maximizing reconstruction distortion. Their results demonstrate improved privacy with minimal impact to task performance. 

Despite the successful results achieved by the aforementioned task-oriented communication systems, there remain several key areas that warrant further investigation and improvement. One significant limitation in the existing literature is the lack of results and comprehensive benchmarking on the mutual information between the encoded features and the input, given a particular task-oriented communication system. In task-oriented communication systems, mutual information plays a critical role in determining the efficiency of the system, particularly regarding the preservation of information during transmission.

Furthermore, while task-specific performance, specifically classification accuracy, has been improved by these advancements, there is still room for enhancing performance further.  Another key challenge in these systems is balancing multiple trade-offs, such as task performance, informativeness, minimality and privacy. These trade-offs are typically managed through careful tuning of hyperparameters. Reducing the dependency on hyperparameters would make these systems more robust and easier to deploy in real-world scenarios.

Contrastive learning has gained significant attention in recent years, particularly for its success in unsupervised learning \cite{oord2018representation, 10.5555/3524938.3525087, tian2020makes, chen2020big}. By leveraging the concept of instance discrimination, contrastive learning methods aim to pull together positive pairs for example, augmentations of the same image, while pushing apart negative pairs for example, different images, thus learning meaningful representations of data without the need for labels. More recently, contrastive learning has also shown exceptional results in supervised learning scenarios. By incorporating label information into the contrastive loss function, methods such as supervised contrastive learning (SupCon) \cite{khosla2020supervised} have improved upon traditional cross-entropy loss. Contrastive learning has not been investigated for task-oriented communication systems as an alternative to cross-entropy based mutual information approximation techniques.

Disentangled representation learning has been widely studied in recent years.  Prominent examples include the $\beta$-VAE \cite{higgins2017beta}, that extends the variational autoencoder (VAE) by introducing a regularization term that encourages disentanglement, and FactorVAE \cite{kim2018disentangling}, that further improves this disentanglement by encouraging the representations' distribution  to be factorial and therefore independent across the dimensions. Other works \cite{sanchez2020learning, pan2021disentangled, 10547074}, explored disentangling through adversarial-based objectives \cite{goodfellow2014generative}.

\section{System Model and Notations}

\subsection{Notations}
Throughout this paper, we use the following notational conventions. Random variables are denoted by uppercase letters, such as $X$, $Y$, and $Z$. Their corresponding realizations (i.e., specific instances) are denoted by lowercase bold letters, such as $\boldsymbol{x}$, $\boldsymbol{y}$, and $\boldsymbol{z}$. The space from which these random variables are drawn is represented by calligraphic letters, such as $\mathcal{X}$, $\mathcal{Y}$  and $\mathcal{Z}$. We denote entropy of a random variable $X$ by $H(X)$. The mutual information between two random variables $X$ and $Y$, is denoted by $I(X; Y)$. We use $I(Z; X | Y)$ to denote the conditional mutual information between $Z$ and $X$ given $Y$. We use the expectation notation \( \mathbb{E}[\cdot] \), which refers to the average value of a random variable over a distribution. Most of the symbols in the article are listed in Table \ref{tab:symbols}.

\subsection{System Model}

We consider a semantic communication system designed for next-generation 6G-enabled Internet of Things (6G-IoT) networks, where distributed edge devices must transmit task-relevant information to centralized or cloud-based servers under strict constraints on bandwidth, latency, and privacy. In such settings, communication should prioritize the efficient delivery of minimal yet informative representations, discarding irrelevant or sensitive content that is unnecessary for the downstream task.

The transmitter includes a feature extractor and a joint source-channel coding (JSCC) encoder. We collectively refer to these components as the task-relevant encoder. The task-relevant encoder encodes an input image $\boldsymbol{x} \in \mathcal{X}$ into a lower-dimensional feature vector $\boldsymbol{z} \in \mathcal{Z}$. Encoded vector $\boldsymbol{z}$ is then transmitted to a receiver over a noisy wireless channel. The primary objective is to transmit a minimal and informative representation, by discarding task-irrelevant information while ensuring that $\boldsymbol{z}$ contains only the essential information for accurate downstream classification at the receiver.

The overall transmission and decoding process can be described by the following Markov chain:
\begin{equation}
    Y \to X \to Z \to \hat{Z} \to \hat{Y},
\end{equation}
where $X$ is the random variable representing the input images, $Z$ is the random variable representing the encoded feature vectors, $\hat{Z}$ is the noisy signals received by the receiver, $Y$ is the random variable representing the labels of the input images, and $\hat{Y}$ is the random variable representing the predicted labels at the receiver.

At the transmitter, the task-relevant encoder encodes input image \( \boldsymbol{x} \in \mathbb{R}^{N} \), where \( N \) represents the number of pixels in the image (Height \(\times\) Width \(\times\) Color Channels). The encoder maps this input into a lower-dimensional feature vector \( \boldsymbol{z} \in \mathbb{R}^{d} \), where \( d \) is the dimension of the encoded feature vector. The encoding function, denoted by \( f_\theta: \mathbb{R}^{N} \to \mathbb{R}^{d} \), is parameterized by \( \theta \), and the encoding process can be expressed as
\begin{equation}
  \boldsymbol{z} = f_\theta(\boldsymbol{x})  
\end{equation}

Feature vector \( \boldsymbol{z} \) is then prepared for transmission over the wireless channel by being mapped to channel input symbols. The role of the task-relevant encoder is twofold: encoding the input data into feature representations and preparing them as channel symbols suitable for transmission. The encoded feature vector \( \boldsymbol{z} \in \mathbb{R}^{d} \) is transmitted over a wireless channel, which is modeled as an additive white Gaussian noise (AWGN) channel. The channel introduces noise and distortion, and the received signal \( \hat{\boldsymbol{z}} \in \mathbb{R}^{d} \) at the receiver is expressed as
\begin{equation}
    \hat{\boldsymbol{z}} = \boldsymbol{z} + \boldsymbol{n},
\end{equation}
where \( \boldsymbol{n} \sim \mathcal{N}(0, \sigma^2 \mathbf{I}) \) is the additive Gaussian noise with variance \( \sigma^2 \). The noise variance \( \sigma^2 \) is related to the channel's \textit{signal-to-noise ratio} (SNR), which quantifies the channel quality. The SNR in decibels (dB) is given by

\begin{equation}
    \text{SNR}_{\text{dB}} = 10 \log_{10}\left(\frac{\mathbb{E}[\|\boldsymbol{z}\|^2]}{\sigma^2}\right).
\end{equation}

At the receiver, typically a cloud or edge server, a classifier, denoted by \( q_\phi: \mathbb{R}^{d} \to \mathbb{R}^{M} \), where \( M \) is the number of labels, is parameterized by \( \phi \). The classifier maps the received noisy signal \( \hat{\boldsymbol{z}} \) to predicted label \( \hat{y} \in \mathbb{R}^{M} \) as

\begin{equation}
   \hat{\boldsymbol{y}} = q_\phi(\hat{\boldsymbol{z}}). 
\end{equation}
The classifier is trained to minimize the loss between the predicted label (\( \hat{\boldsymbol{y}} \)) and the true label (\( \boldsymbol{y} \)). Since true posterior distribution \( p(\boldsymbol{y} | \hat{\boldsymbol{z}}) \) is intractable, \( q_\phi \) serves as an approximation based on the received noisy signal.

\begin{table}[!t]
\centering
\caption{Description of Symbols}
\label{tab:symbols}
\renewcommand{\arraystretch}{1.2}
\begin{tabular}{ll}
\toprule
\textbf{Symbol} & \textbf{Description} \\
\midrule
$\boldsymbol{y}, Y$ & Target variable and its realization \\
$\boldsymbol{x}, X$ & Input variable and its realization \\
$\boldsymbol{z}, Z$ & Encoded feature and its realization \\
$\hat{\mathbf{z}}, \hat{Z}$ & Received (noisy) encoded feature and its realization \\
$\boldsymbol{z}_1, Z_1$ & Task-relevant feature and its realization \\
$\boldsymbol{z}_2, Z_2$ & Task-irrelevant feature and its realization \\
$\hat{\boldsymbol{y}}, \hat{Y}$ & Predicted label and its realization \\
$\mathcal{X}, \mathcal{Y}, \mathcal{Z}$ & Input, label, and feature spaces \\
$H(\cdot)$ & Entropy function \\
$I(\cdot; \cdot), I(\cdot; \cdot \mid \cdot)$ & Mutual and conditional mutual information \\
$\mathbb{E}[\cdot]$ & Expectation operator \\
$\mathbf{I}$ & Identity matrix \\
$p(\cdot)$ & Probability distribution \\
$\mathcal{N}$ & Statistical Gaussian distribution \\
$\mathbf{n}$ & Additive Gaussian noise \\
$f_\theta(\cdot)$ & Task-relevant encoder, parameterized by $\theta$ \\
$g_\eta(\cdot)$ & Task-irrelevant encoder, parameterized by $\eta$ \\
$r_\omega(\cdot)$ & Reconstructor, parameterized by $\omega$ \\
$h_\psi(\cdot)$ & Projection head , parameterized by $\psi$ \\
$q_\phi(\cdot)$ & Classifier, parameterized by $\phi$ \\
$D_\nu(\cdot)$ & Discriminator, parameterized by $\nu$ \\

\bottomrule

\end{tabular}
\end{table}

\begin{figure}
    \centering
    \includegraphics[width=0.85\linewidth]{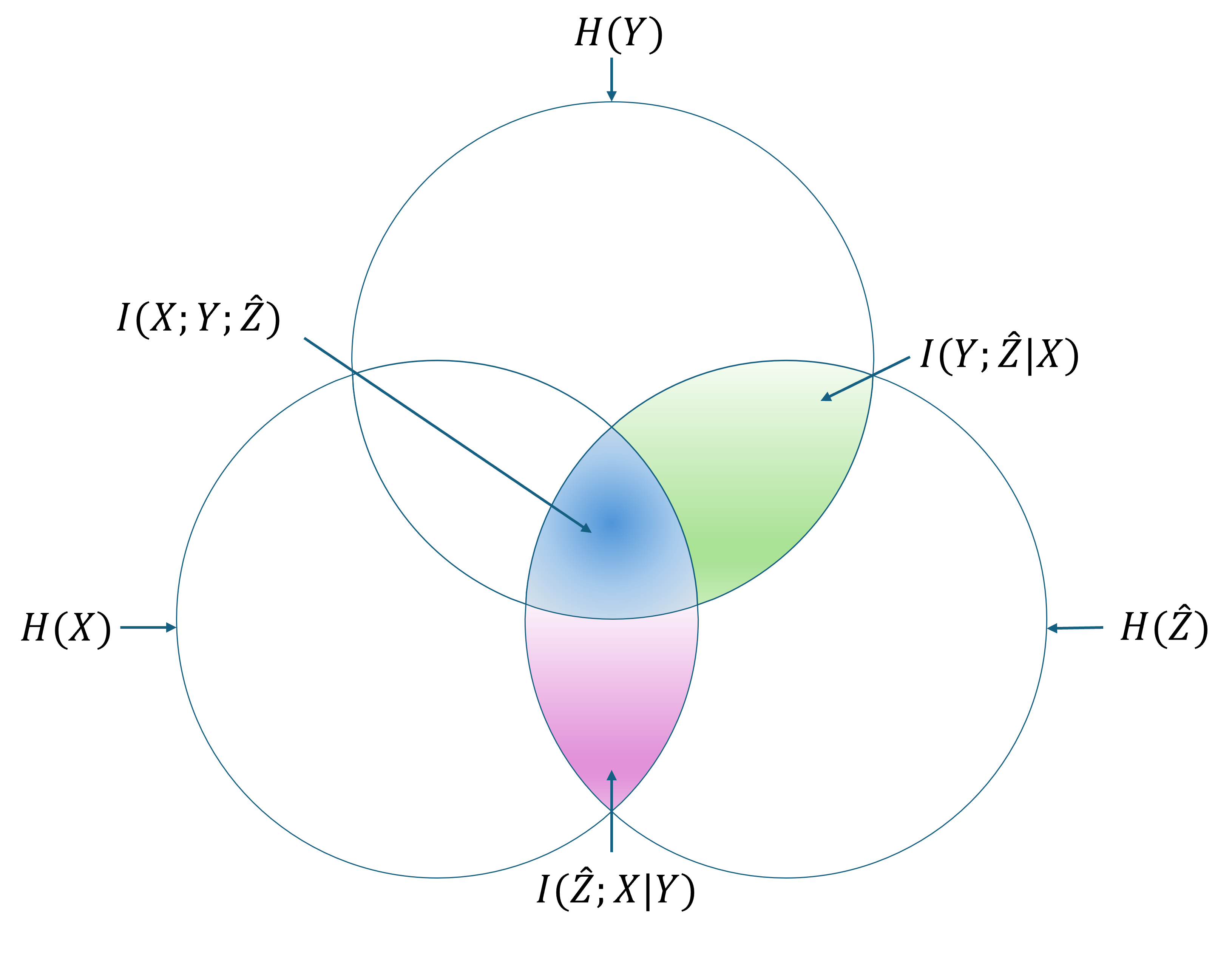}
    \caption{Information diagram for three random variables $X$, $Y$, $\hat{Z}$. The union of the blue and pink regions yields \(I(\hat{Z}; X)\), and the union of the blue and green regions yields \(I(\hat{Z}; Y)\).}
    \label{fig:Venn}
\end{figure}

\section{Problem Description}

In this section, we identify the primary challenges that arise when transmitting features over a communication channel and utilizing them for a downstream classification task. Our goal is to ensure that the transmitted features contain only the minimum necessary information required for the downstream task to maximize efficiency whilst being privacy aware. Furthermore, we argue that it is necessary to have a fair, reproducible, and unified method to obtain comparative values that act as a proxy for mutual information between the encoded features and the input data to allow effective benchmarking of different task-oriented communication systems.

\subsection{Minimum Necessary Information}

Following  \cite{fischer2020conditional}, the Minimum Necessary Information (MNI) criterion for an ideal representation $\hat{Z}$ under ideal transmission conditions must satisfy the following key principles:

\begin{itemize}
    \item \textbf{Informativeness:} Representation $\hat{Z}$ should contain all the necessary information to predict $Y$, requiring us to maximize the mutual information $I(\hat{Z}; Y)$.
    \item \textbf{Necessity:} Representation $\hat{Z}$ should contain the necessary amount of information in order to perform well in the downstream task, any less information would mean that $Z$ has discarded task-relevant information. Necessity can be defined as
\begin{equation}
    I(X; Y) \leq I(Y; \hat{Z})
\end{equation}

    \item \textbf{Minimality:} Among all possible representations $Z$ that satisfy the task of predicting $Y$, we seek the one that encodes the least amount of information about $X$ beyond what is strictly necessary for the task. This can be formulated as
    \begin{equation}
     \min_{\hat{Z}} I(\hat{Z}; X) \quad \text{subject to} \quad I(X; Y) = I(\hat{Z}; Y)       
    \end{equation}

    Any more information than that would result in $Z$ having redundant information about $X$ that is unnecessary for predicting $Y$.

\end{itemize}

Given the above we conclude that in an optimal case under ideal channel conditions we must have
\begin{equation}
    I(\hat{Z}; X) = I(\hat{Z};Y) = I(X;Y),
\end{equation}
this implies that $\hat{Z}$ contains exactly the amount of information necessary to perform the task of predicting $Y$ from $X$, no more and no less. At the MNI point, $\hat{Z}$ captures all the relevant information needed for the task, while discarding any irrelevant or redundant information about $X$.

\subsection{Privacy Concerns and Task-Irrelevant Information}
The second challenge is privacy concerns due to the leakage of task-irrelevant information from $X$ into $\hat{Z}$. If $\hat{Z}$ retains information about $X$ that is not relevant to predicting $Y$, this may inadvertently expose sensitive or private data, and could make the system more vulnerable to different attacks such as attribute inference attacks and  model inversion attacks \cite{shokri2017membership}, \cite{fredrikson2015model}. Therefore, disentangling task-irrelevant information ensures that $\hat{Z}$ does not encode unnecessary or sensitive information that is not directly relevant to the downstream task, which minimizes privacy risks.

\subsection{Quantifying Information Retention}
In task-oriented communication systems, it is critical to have an understanding of the mutual information \(I(\hat{Z}; X)\) as it provides insights into how much of the original input information \(X\) is encoded in \(\hat{Z}\) and can directly affect latency, bandwidth and privacy. However, a significant challenge arises because the estimation of \(I(\hat{Z}; X)\) varies drastically depending on the estimation method used. Indeed, multiple works have reported widely different \(I(\hat{Z}; X)\) for the same task-oriented approach \cite{alemi2017deep}, \cite{pan2021disentangled}. This makes it difficult to arrive at reliable conclusions about the amount of information being retained in \(\hat{Z}\).

Given these discrepancies, we argue that it is crucial to devise a method that yields consistent, reliable and fair comparative estimates of information retention, even if the exact value of \(I(\hat{Z}; X)\) is intractable. Instead of absolute precision, a method that provides relative and comparable estimates across different systems would greatly enhance the ability to evaluate and optimize different task-oriented communication systems.

\subsection{Limitations of Variational Information Bottleneck (VIB)}

The variational information bottleneck (VIB) has been the de facto method for many task-oriented communication systems. VIB tries to minimize the following objective:
\begin{equation}
   \mathcal{L}_{\text{VIB}} = \beta I(\hat{Z}; X) - I(\hat{Z}; Y), 
\end{equation}
where \(I(\hat{Z}; X)\) measures the amount of information retained from the input \(X\), and \(I(\hat{Z}; Y)\) represents the informativeness of \(Z\) for predicting \(Y\). Hyperparameter \( \beta \) controls the trade-off between preserving task-relevant information and discarding irrelevant information. To maximize \(I(\hat{Z}; Y)\), a cross-entropy based loss is used, and the Kullback–Leibler divergence is used to to minimize \(I(\hat{Z}; X)\) \cite{alemi2017deep, shao2021learning, xie2023robust}.

However, VIB-based task-oriented communication systems presents several challenges:
\begin{itemize}
    \item \textbf{Limitation of cross-entropy based loss:} The majority of task-oriented communication systems rely on the cross-entropy loss as a vraitaional approximation to maximize \(I(\hat{Z}; Y)\). However, it has been shown recently that supervised contrastive learning based loss \cite{khosla2020supervised} outperforms the cross-entropy loss in different settings. 
    \item \textbf{Conflicting objectives:} The VIB objective which maximizes \(I(\hat{Z}; X)\) and minimizes  \(I(\hat{Z}; Y)\) leads to a conflicting objective as shown by \cite{fischer2020conditional}. If we consider the information diagram \cite{infodiagram} presented in Fig. \ref{fig:Venn}, it is evident that the region shaded in blue, namely \(I(X; Y; \hat{Z})\)  is a subset of \(I(\hat{Z}; X)\) and \(I(\hat{Z}; Y)\)  and is therefore maximized and minimized simultaneously.
    \item \textbf{Inadequate disentanglement:} VIB does not explicitly enforce a separation between the portions of \( Z \) that are relevant for predicting \( Y \) and those that capture irrelevant or redundant details from \( X \). This lack of disentanglement can compromise the privacy and efficiency of the transmitted representation.
\end{itemize}

\section{Proposed Method: CLAD}
The core idea of CLAD is to create an end-to-end communication system that explicitly learns to disentangle task-relevant information from task-irrelevant content, enabling both high task accuracy and improved privacy. CLAD integrates supervised contrastive learning, adversarial training, and reconstruction-based supervision into a unified, structured framework that aligns with the MNI principle.

Each component in CLAD is designed to target a specific challenge:
\begin{itemize}
    \item \textbf{Contrastive learning} maximizes the informativeness of the representation with respect to the downstream task.
    \item \textbf{Adversarial disentanglement} promotes independence between task-relevant and irrelevant features, suppressing information leakage.
    \item \textbf{Reconstruction learning} ensures that task-irrelevant features are adequately captured.
\end{itemize}

To ensure effective optimization, CLAD adopts a \textit{three-stage training strategy}, where each stage isolates and refines a different component of the system. This design is intentional as it prevents interference between competing objectives, stabilizes training, and encourages modular reuse of the encoders across stages.

\begin{figure}
    \centering
    \includegraphics[width=\linewidth]{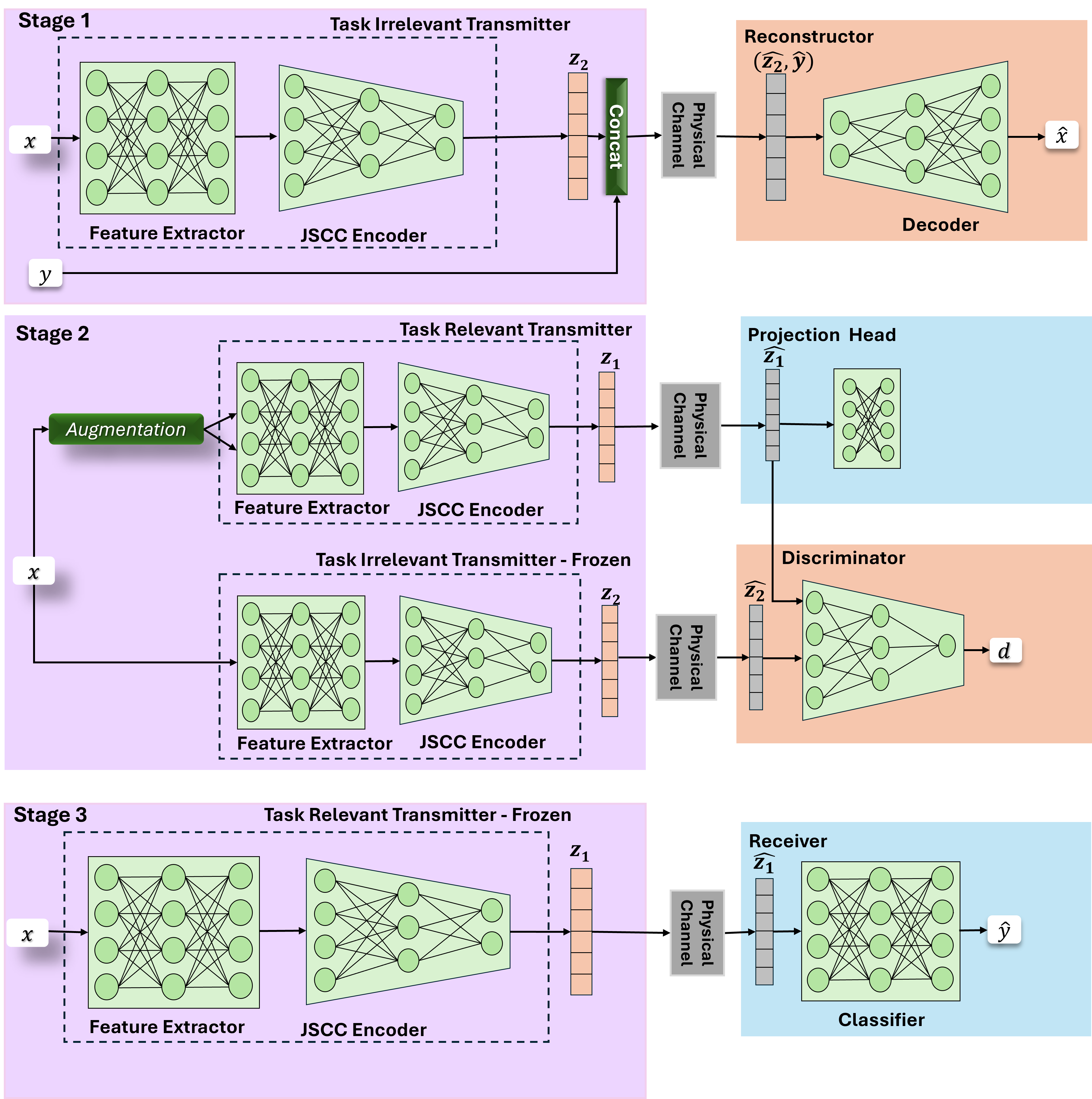}
    \caption{Three stages for training CLAD}
    \label{fig:CLAD}
\end{figure}

CLAD utilizes two key encoders: the  task-relevant encoder, which maps the input into the task-relevant channel codeword \( Z_1 \), and the task-irrelevant encoder , which maps the input into task-irrelevant channel  codeword \( Z_2 \). The task performance is optimized through contrastive learning, which aims to maximize the mutual information \( I(\hat{Z_1}; Y) \), ensuring that \( \hat{Z_1} \) captures the most informative features for downstream classification. The disentanglement is achieved through reconstruction learning to capture task-irrelevant information in \( \hat{Z_2} \), and adversarial training is utilized to minimize the mutual information \( I(\hat{Z_1}; \hat{Z_2)} \), thus promoting independence between the two feature representations. The different components and training stages of CLAD are visualized in Fig. \ref{fig:CLAD}.  These components are optimized together to ensure both high task accuracy and effective disentanglement of information, corresponding to the following maximization objective:

\begin{align}
    \mathcal{L}_{\text{CLAD}} = I(\hat{Z_1}; Y) + I(\hat{Z_2}; X | Y) - I(\hat{Z_1}; \hat{Z_2}).
\end{align}

Here, the objective consists of three key terms:
\begin{itemize}
    \item $I(\hat{Z_1}; Y)$ maximizes the mutual information between task-relevant features $\hat{Z_1}$ and the label $Y$ using contrastive learning;
    \item $I(\hat{Z_2}; X | Y)$ ensures that $\hat{Z_2}$ captures the residual information in $X$ that is not covered by $Y$ by utilizing a reconstruction loss;
    \item $I(\hat{Z_1}; \hat{Z_2})$ minimizes the information overlap between $\hat{Z_1}$ and $\hat{Z_2}$, encouraging disentanglement via an adversarial loss.
\end{itemize}

Reflecting back on Fig. \ref{fig:Venn}, we can see that our new objective maximizes the blue region (task-relevant information) and minimizes the pink region (task-irrelevant information) and avoids the conflicting objectives of VIB.
We explain each of the components of our loss function in detail below, accompanied by their mathematical formulations, implementation details, and training strategy. 

\begin{figure*}[t]
    \centering
    \subfloat[]{
        \centering
        \includegraphics[width=0.3\linewidth]{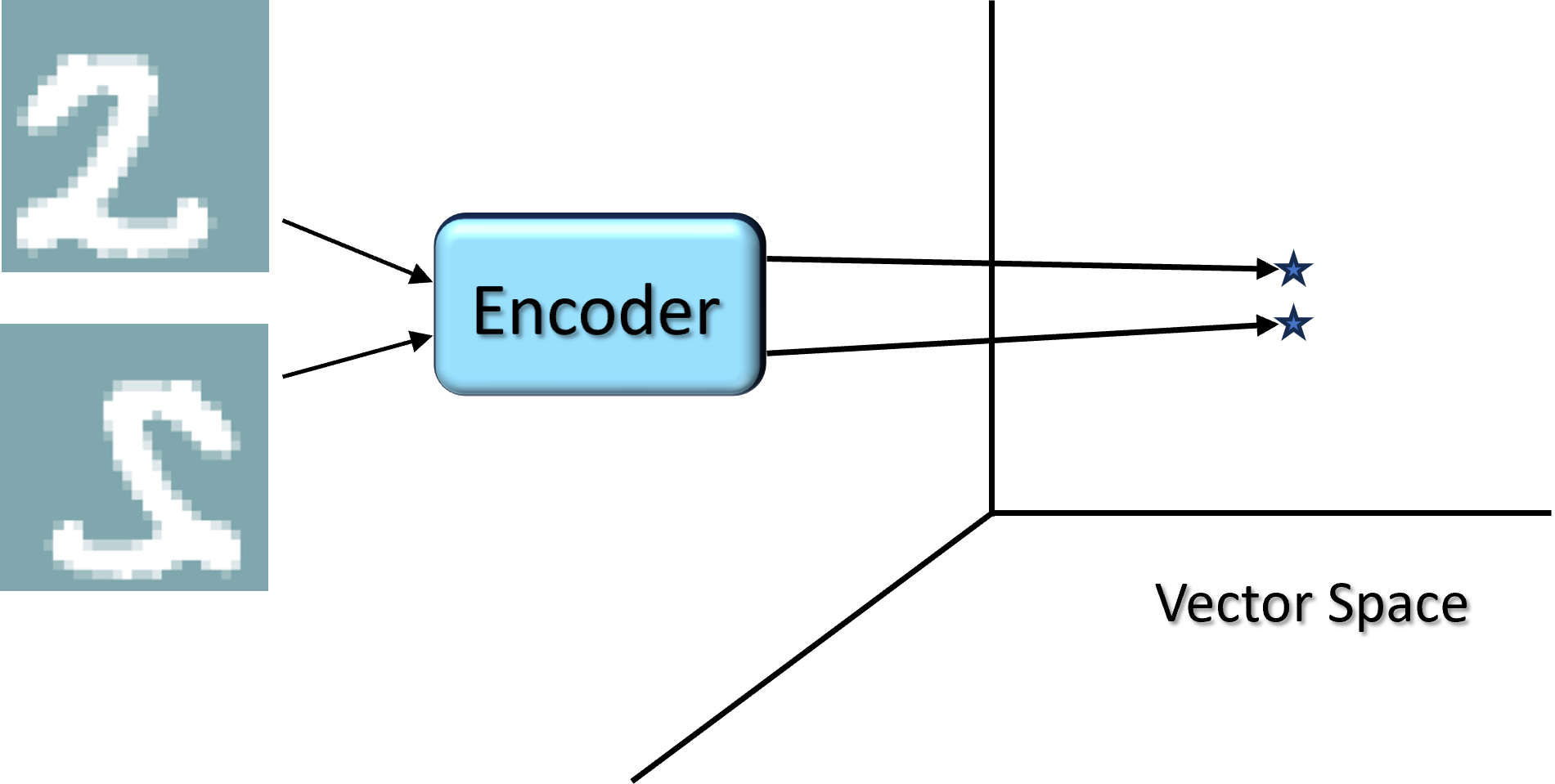}
        \label{subfig:illu_tradeoff_1}
    }
    \subfloat[]{
        \centering
        \includegraphics[width=0.3\linewidth]{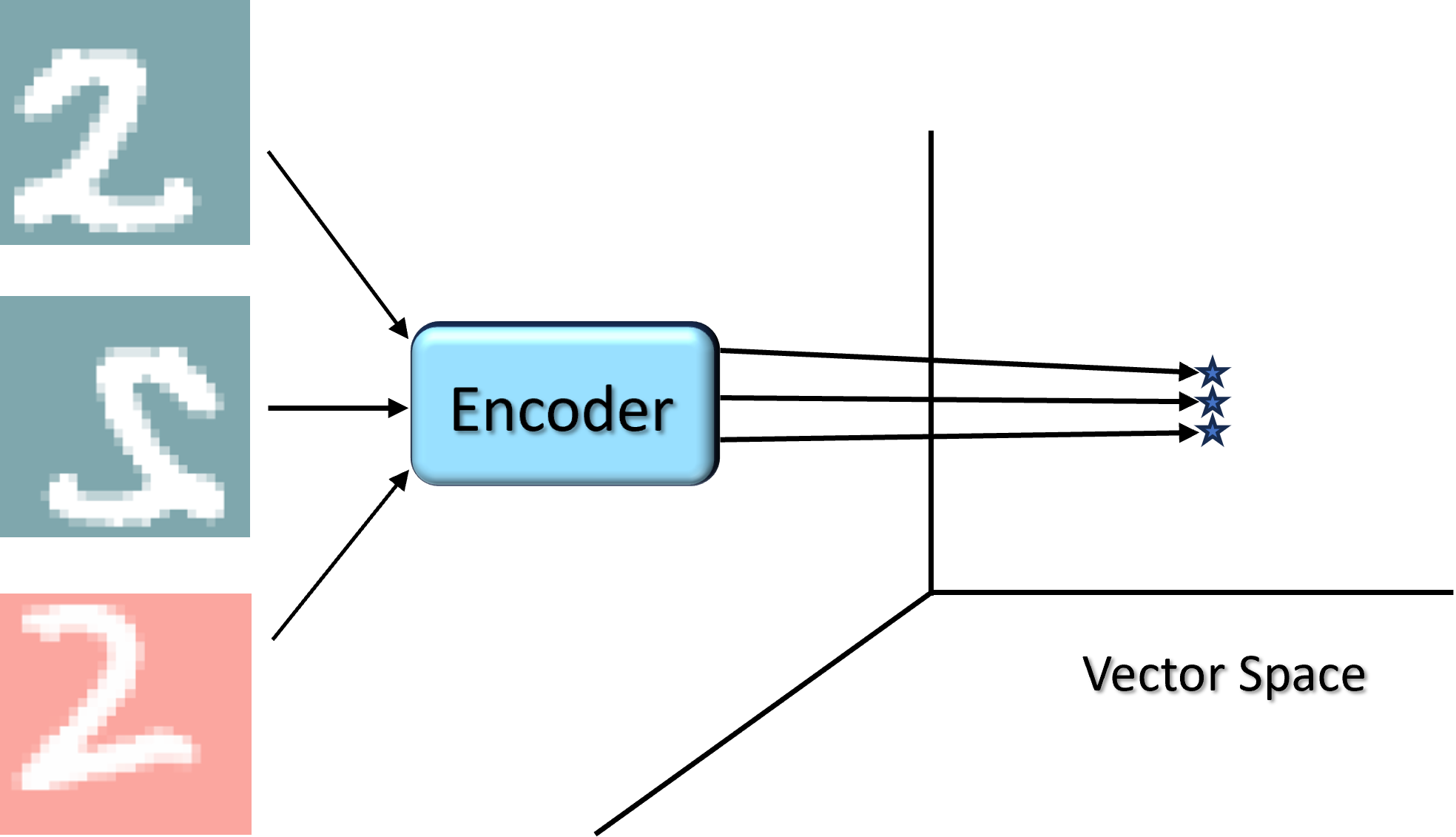}
        \label{subfig:illu_tradeoff_2}
    }
    \subfloat[]{
        \centering
        \includegraphics[width=0.3\linewidth]{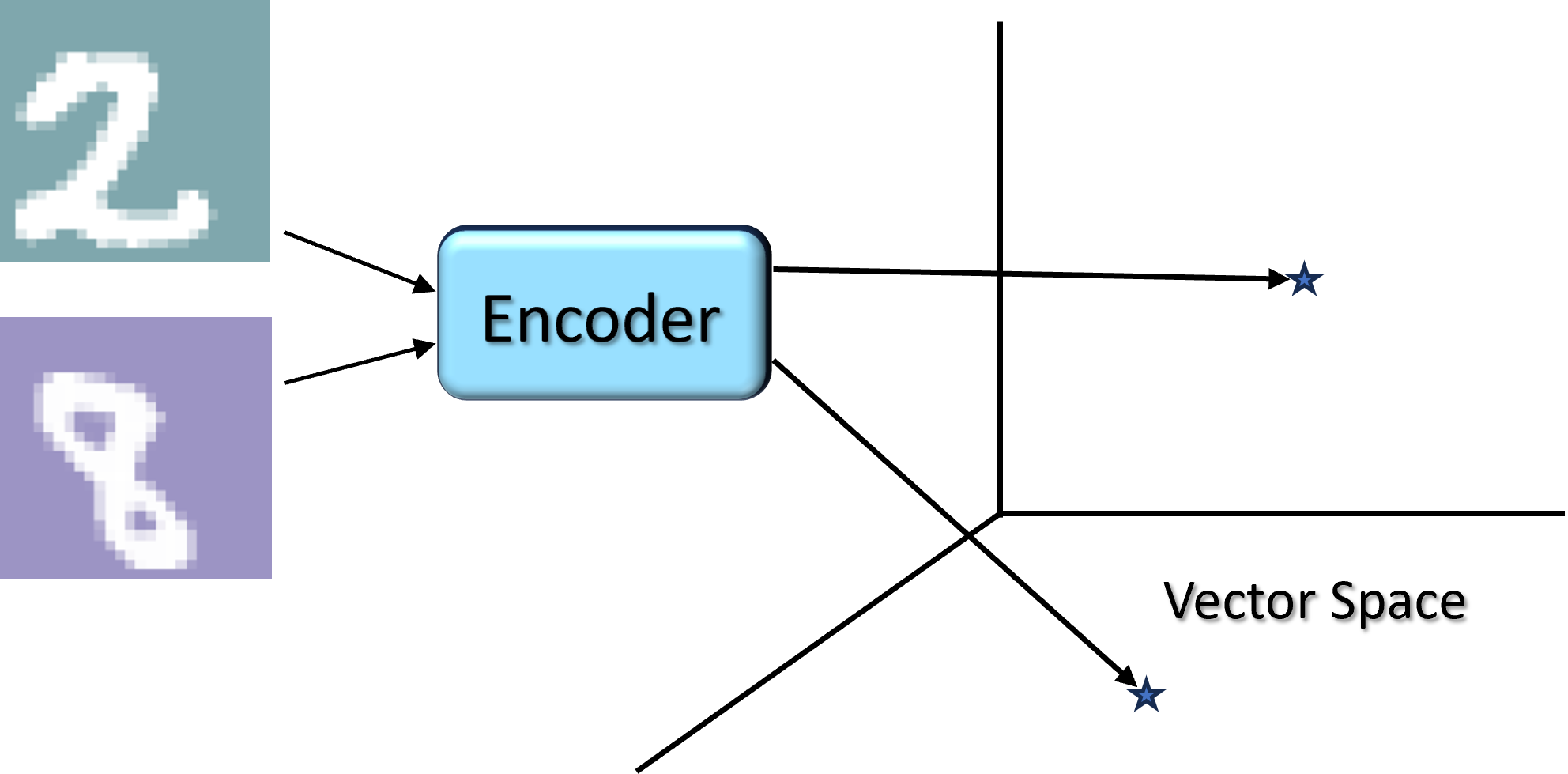}
    }
    \caption{(a): Self-supervised contrastive learning: The model works only on augmentations of the same image; (b): Supervised contrastive learning: Label information is used to align similar classes in vector space; and (c) Both self-supervised and supervised contrastive learning push apart different images in the vector space.}
    \label{fig:contrastive}
\end{figure*}

\begin{algorithm}[]
\caption{Stage 1: Train Task-Irrelevant Encoder with Reconstructor}
\label{alg:train_encoder_reconstructor}
\begin{algorithmic}[1]
\Require $\mathcal{X}_{train}$ (Training dataset), $\kappa$ (SNR), $\lambda$ (Learning rate)
\Ensure Frozen task-irrelevant encoder $g_\eta$
\State Initialize $\eta$, $\omega$
\While{not converged}
    \State Sample $(\boldsymbol{x}, \boldsymbol{y}) \sim \mathcal{X}_{train}$
    \State $\boldsymbol{z}_2 \leftarrow g_\eta(\boldsymbol{x})$
    \State $\boldsymbol{n} \sim \mathcal{N}(0, \sigma^2 \mathbf{I})$, $\sigma^2 \leftarrow \frac{\mathbb{E}[\|\boldsymbol{z}_2\|^2]}{10^{\frac{\kappa}{10}}}$
    \State $\hat{\boldsymbol{z}}_2 \leftarrow \boldsymbol{z}_2 + \boldsymbol{n}$
    \State $\hat{\boldsymbol{x}} \leftarrow r_\omega(\hat{\boldsymbol{z}}_2, \boldsymbol{y})$
    \State $\mathcal{L}_{\text{recon}} \leftarrow \| \boldsymbol{x} - \hat{\boldsymbol{x}} \|^2$
    \State $\eta \leftarrow \eta - \lambda \nabla_{\eta} \mathcal{L}_{\text{recon}}$
    \State $\omega \leftarrow \omega - \lambda \nabla_{\omega} \mathcal{L}_{\text{recon}}$
\EndWhile
\State Discard $r_\omega$
\State Freeze $\eta$
\end{algorithmic}
\end{algorithm}

\subsection{Contrastive Loss for Task-Relevant Features}

To maximize \( I(\hat{Z}_1; Y) \), we adopt a supervised contrastive learning framework similar to \cite{khosla2020supervised}. First, we apply an augmentation function, which applies different augmentations such as cropping, rotating and reflecting, \( \texttt{Aug}(\cdot) \) to the input image \( \boldsymbol{x} \) to generate two different views \( \tilde{\boldsymbol{x}}_1 = \texttt{Aug}(\boldsymbol{x}) \) and \( \tilde{\boldsymbol{x}}_2 = \texttt{Aug}(\boldsymbol{x}) \). These augmented samples are then passed through the TRE, denoted by \( f_\theta(\cdot) \), resulting in two representations, \( \boldsymbol{z}_1 = f_\theta(\tilde{\boldsymbol{x}}_1) \) and \( \boldsymbol{z}_2 = f_\theta(\tilde{\boldsymbol{x}}_2) \), where \( \boldsymbol{z} \in \mathbb{R}^{d} \).

Following that, these representations are transmitted over the physical channel and projected into a lower-dimensional space through a projection head, \( h_\psi(\cdot) \) parameterized by \( \psi\), yielding \( \boldsymbol{h}_1 = h_\psi(\boldsymbol{\hat{z}}_1) \) and \( \boldsymbol{h}_2 = h_\psi(\boldsymbol{\hat{z}}_2) \), where \( \boldsymbol{h} \in \mathbb{R}^{d_p} \), and \( d_p \) represents the dimensionality of the projection space.

A contrastive loss, \( \mathcal{L}_{\text{contrast}} \), is designed to maximize the agreement between representations of similar class samples while minimizing the similarity between representations of different class samples. Considering a batch of intermediate features \([ \boldsymbol{z}_1, \dots, \boldsymbol{z}_B ]\) and their corresponding labels \([\boldsymbol{y}_1, \dots, \boldsymbol{y}_B ]\), the loss function is defined as
\begin{equation}
    S_{ij} = \frac{\exp\left(\boldsymbol{h}_i^\top \boldsymbol{h}_j / \tau\right)}{\sum_{k=1}^{B} \mathds{1}_{i \neq k} \exp\left(\boldsymbol{h}_i^\top \boldsymbol{h}_k / \tau\right)},
\end{equation}
\begin{equation}
    \mathcal{L}_{\text{contrast}} = -\frac{1}{\sum_{i \neq j} \mathds{1}_{y_i=y_j}} \sum_{i \neq j} \mathds{1}_{y_i=y_j} \log S_{ij},
\end{equation}
where \( S_{ij} \) represents the similarity score between the projected representations \( \boldsymbol{h}_i \) and \( \boldsymbol{h}_j \), \( \tau \) is a temperature scaling factor, and \( \mathds{1}_{y_i=y_j} \) is an indicator function that equals 1 if \( y_i = y_j \) (positive pairs) and 0 otherwise (negative pairs). This loss encourages encoder \( f_\theta(\cdot) \)  to learn class-discriminative features, ensuring that the latent representation \( \boldsymbol{z} \) captures the necessary information for the downstream classification task. Supervised contrastive learning is presented visually in Fig \ref{fig:contrastive}.

Next, we prove that minimizing the contrastive loss, defined above, maximizes a lower bound on the task-relevant information \( I({\hat{Z}}; Y) \). We begin by considering a simplified situation where we have a query sample \(\boldsymbol{h}^{+}\) together with a set \(\boldsymbol{H} = \{\boldsymbol{h}_1, \dots, \boldsymbol{h}_B\}\) consisting of \(B\) samples. In this set, one sample \(\boldsymbol{h}^{p}\) is a positive sample from the same class as \(\boldsymbol{h}^{+}\), while the other negative samples are randomly sampled. Namely, \(\boldsymbol{H} = \{\boldsymbol{h}^{p}\} \cup \boldsymbol{H}_{\text{neg}}\). The expectation of the contrastive loss is given by

\begin{equation}
   \label{eq:ec}
   \mathbb{E}[\mathcal{L}_{\text{contrast}}] = \mathbb{E}_{\boldsymbol{h}^{+}, \boldsymbol{H}}\left[
      - \log \frac{\exp({\boldsymbol{h}^{+}}^{\top} \boldsymbol{h}^{p} / \tau)}{
         \sum_{i=1}^{B} \exp({\boldsymbol{h}^{+}}^{\top} \boldsymbol{h}_i  / \tau)
     } \right].
\end{equation}

Equation (\ref{eq:ec}) can be viewed as a categorical cross-entropy loss for recognizing the positive sample \(\boldsymbol{h}^{p}\). We define the optimal probability of identifying the positive sample as

\begin{equation}
   P(\boldsymbol{h}_i|\boldsymbol{H}) = \frac{p(\boldsymbol{h}_i|y) \prod_{l\neq i} p(\boldsymbol{h}_l)}{\sum_{j=1}^{B} p(\boldsymbol{h}_j|y) \prod_{l\neq j} p(\boldsymbol{h}_l)}
   =\frac{ \frac{p(\boldsymbol{h}_i|y)}{p(\boldsymbol{h}_i)} }{\sum_{j=1}^{B} \frac{p(\boldsymbol{h}_j|y)}{p(\boldsymbol{h}_j)}}.
\end{equation}
This shows that the optimal value of \(\exp({\boldsymbol{h}^{+}}^{\top} \boldsymbol{h}^{p} / \tau)\) is \(\frac{p(\boldsymbol{h}^{p}|y)}{p(\boldsymbol{h}^{p})}\). Assuming that \(\boldsymbol{h}^{+}\) is uniformly sampled from all classes, we derive the following bound,

\begin{align}
   \mathbb{E}[\mathcal{L}_{\text{contrast}}] &\geq
   \mathbb{E}[\mathcal{L}_{\text{contrast}}^{\text{optimal}}] \nonumber \\
   &= \mathbb{E}_{y, \boldsymbol{H}}
   \left[
   - \log \frac{ \frac{p(\boldsymbol{h}^{p}|y)}{p(\boldsymbol{h}^{p})} }{\sum_{j=1}^{B} \frac{p(\boldsymbol{h}_j|y)}{p(\boldsymbol{h}_j)}} \right] \\
   &= \mathbb{E}_{y, \boldsymbol{H}}
   \left[
   - \log \frac{ \frac{p(\boldsymbol{h}^{p}|y)}{p(\boldsymbol{h}^{p})} }{
      \frac{p(\boldsymbol{h}^{p}|y)}{p(\boldsymbol{h}^{p})} +
   \sum_{\boldsymbol{h}_j \in \boldsymbol{H}_{\text{neg}}} \frac{p(\boldsymbol{h}_j|y)}{p(\boldsymbol{h}_j)}} \right] \nonumber \\
   &= \mathbb{E}_{y, \boldsymbol{H}}\left\{
   \log \left[ 1 + \frac{p(\boldsymbol{h}^{p})}{p(\boldsymbol{h}^{p}|y)} 
   \sum_{\boldsymbol{h}_j \in \boldsymbol{H}_{\text{neg}}} \frac{p(\boldsymbol{h}_j|y)}{p(\boldsymbol{h}_j)} \right]\right\}.
\end{align}
For large \(B\), from the law of large numbers, we can approximate the sum of negative samples by its expected value as follows,

\begin{align}
   \label{eq:approx}
   \approx \mathbb{E}_{y, \boldsymbol{H}}\left\{
   \log \left[ 1 + \frac{p(\boldsymbol{h}^{p})}{p(\boldsymbol{h}^{p}|y)} 
   (B-1) \mathbb{E}_{\boldsymbol{h}_j \sim p(\boldsymbol{h}_j)} \frac{p(\boldsymbol{h}_j|y)}{p(\boldsymbol{h}_j)} \right]\right\}.
\end{align}
Since the negative samples are class-neutral (i.e., independent of \( y \)), the inner expectation over negative samples \( \boldsymbol{h}_j \) simplifies to a constant value. This allows us to focus the outer expectation on \( y \) and \( \boldsymbol{h}^p \), concentrating on the probability of correctly identifying the positive sample among the negatives as follows,

\begin{align}
   = \mathbb{E}_{y, \boldsymbol{h}^{p}} &\left\{
      \log \left[ 1 + \frac{p(\boldsymbol{h}^{p})}{p(\boldsymbol{h}^{p}|y)} 
      (B-1) \right] \right\} \nonumber \\
   &\geq \mathbb{E}_{y, \boldsymbol{h}^{p}} \left\{
      \log \left[ \frac{p(\boldsymbol{h}^{p})}{p(\boldsymbol{h}^{p}|y)} 
      (B-1) \right] \right\} \nonumber \\
   &= \mathbb{E}_{y, \boldsymbol{h}^{p}} \left\{
      - \log \left[ \frac{p(\boldsymbol{h}^{p}|y)}{p(\boldsymbol{h}^{p})} \right]
      + \log(B-1) \right\} \nonumber \\
   &= -I(\boldsymbol{h}^{p}; y) + \log(B-1) \geq -I(\boldsymbol{\hat{z}}; y) + \log(B-1).
   \label{eq:final_approx}
\end{align}
From the above, the last inequality in (\ref{eq:final_approx}) follows from the data processing inequality \cite{tishby2000information}. Finally, we conclude that

\begin{equation}
  \mathbb{E}[\mathcal{L}_{\text{contrast}}] \geq \log(B-1) - I(\hat{Z}; Y),
\end{equation}
and thus minimizing \(\mathcal{L}_{\text{contrast}}\) maximizes a lower bound of \( I(\hat{Z}, Y) \). Increasing \( B \) raises \(\log(B-1)\), thereby strengthening this lower bound and enhancing performance by preserving more task-relevant information.  Although the derived bound can be loose with a small number of negative samples~\cite{oord2018representation}, we mitigate this by sampling large batches (2048) during contrastive training.

\subsection{Reconstruction for Task-Irrelevant Features}

To maximize \(I(\hat{Z}_2; X | Y)\), we use a reconstruction-based objective that ensures \(X\) is reconstructed from both \(Y\) and \(\hat{Z}_2\), where \(\hat{Z}_2\) captures the information in \(X\) that is not already captured by \(Y\). Let \( g_\eta: \mathbb{R}^N \to \mathbb{R}^d \) represent the task-irrelevant encoder, parameterized by \( \eta \), which maps input \( \boldsymbol{x} \in \mathbb{R}^N \) to encoded task-irrelevant representation \( \hat{\boldsymbol{z}}_2 \in \mathbb{R}^d \). The encoder approximates the posterior distribution of the latent variable \( \hat{\boldsymbol{z}}_2 \) given \( \boldsymbol{x} \), which we denote by \( q(\hat{\boldsymbol{z}}_2 | \boldsymbol{x}) \). This encoder is responsible for capturing features unrelated to the task, i.e., the features not directly useful for predicting \( \boldsymbol{y} \). Mathematically, the encoded task-irrelevant representation is given by
\begin{equation}
    \boldsymbol{z}_2 = g_\eta(\boldsymbol{x}),
\end{equation}
where \( d \) represents the dimensionality of the task-irrelevant feature space.

Next, we introduce the reconstructor \( r_\omega: \mathbb{R}^d \times \mathbb{R}^M \to \mathbb{R}^N \), parameterized by \( \omega \). The reconstructor \( r_\omega \) takes as input both noisy task-irrelevant features \( \hat{\boldsymbol{z}}_2 \) and task-relevant label \( \boldsymbol{y} \) and attempts to reconstruct the original input \( \boldsymbol{x} \). The objective is to minimize the reconstruction error, ensuring that \( \hat{\boldsymbol{z}}_2 \) focuses solely on task-irrelevant information. The reconstruction loss is defined as 
\begin{equation}
   \mathcal{L}_{\text{recon}} = \mathbb{E}_{p(\boldsymbol{x}, \boldsymbol{y})} \left[ \| r_\omega(\hat{\boldsymbol{z}}_2, \boldsymbol{y}) - \boldsymbol{x} \|^2 \right].
\end{equation}
To justify this approach, we show how this reconstruction-based loss provides an approximation for the mutual information \( I(\hat{Z}_2; X | Y) \). Using a variational encoder and reconstruction model \( r_\omega(\boldsymbol{x} | \hat{\boldsymbol{z}}_2, \boldsymbol{y}) \), we can approximate \( I(\hat{Z}_2; X | Y) \) as follows,

\begin{align}
    I(\hat{Z}_2; X | Y) &\geq \mathbb{E}_{p(\boldsymbol{x}, \boldsymbol{y}) q(\hat{\boldsymbol{z}}_2 | \boldsymbol{x})} 
    \left[ \log p(\boldsymbol{x} | \boldsymbol{y}, \hat{\boldsymbol{z}}_2) \right] \nonumber \\
    &\quad - \mathbb{E}_{p(\boldsymbol{x}, \boldsymbol{y})} \left[ \log p(\boldsymbol{x} | \boldsymbol{y}) \right].
\end{align}
The first term, \( \mathbb{E}_{p(\boldsymbol{x}, \boldsymbol{y}) q(\hat{\boldsymbol{z}}_2 | \boldsymbol{x})} 
    [ \log p(\boldsymbol{x} | \boldsymbol{y}, \hat{\boldsymbol{z}}_2)\), represents the expected log-likelihood of reconstructing \( \boldsymbol{x} \) given both \( \boldsymbol{y} \) and \( \hat{\boldsymbol{z}}_2 \). The second term, \( \mathbb{E}_{p(\boldsymbol{x}, \boldsymbol{y})} \left[ \log p(\boldsymbol{x} | \boldsymbol{y}) \right] \), represents the expected log-likelihood of reconstructing \( \boldsymbol{x} \) based solely on \( \boldsymbol{y} \), independent of the task-irrelevant features.

Minimizing the reconstruction loss \( \mathcal{L}_{\text{recon}} \) effectively approximates the maximization of the first term in the mutual information expression, thereby increasing \( I(\hat{Z}_2; X | Y) \). By optimizing both the encoder \( g_\eta \) and the reconstructor \( r_\omega \), we ensure that \( \hat{\boldsymbol{z}}_2 \) captures task-irrelevant information while leveraging \( \boldsymbol{y} \) for the reconstruction of task-relevant features in \( \boldsymbol{x} \).

\subsection{Adversarial Disentanglement}

To approximate the minimization of the mutual information \(I(\hat{Z}_1; \hat{Z}_2)\), we employ adversarial training, following the approach in \cite{sanchez2020learning}. This ensures that task-relevant features in \( \hat{Z}_1 \) and task-irrelevant features in \( \hat{Z}_2 \) are disentangled.
The mutual information \(I(\hat{Z}_1; \hat{Z}_2)\) quantifies the dependence between \( \hat{Z}_1 \) and \( \hat{Z}_2 \). It is formally defined as

\begin{equation}
  I(\hat{Z}_1; \hat{Z}_2) = \int_{\hat{\boldsymbol{z}}_1} \int_{\hat{\boldsymbol{z}}_2} p(\hat{\boldsymbol{z}}_1, \hat{\boldsymbol{z}}_2) \log \left( \frac{p(\hat{\boldsymbol{z}}_1, \hat{\boldsymbol{z}}_2)}{p(\hat{\boldsymbol{z}}_1) p(\hat{\boldsymbol{z}}_2)} \right) d\hat{\boldsymbol{z}}_1 d\hat{\boldsymbol{z}}_2.  
\end{equation}
Minimizing it promotes independence between these two representations. However, directly computing \(I(\hat{Z}_1; \hat{Z}_2)\) is intractable since it requires access to the underlying joint distribution \( p(\hat{\boldsymbol{z}}_1, \hat{\boldsymbol{z}}_2)\) and the product of the marginals \( p(\hat{\boldsymbol{z}}_1)p(\hat{\boldsymbol{z}}_2)\). To circumvent this, we approximate the minimization using a discriminator to distinguish between samples drawn from the joint distribution \( p(\hat{\boldsymbol{z}}_1, \hat{\boldsymbol{z}}_2)\) and samples drawn from the product of the marginals \( p(\hat{\boldsymbol{z}}_1)p(\hat{\boldsymbol{z}}_2)\).

To approximate the joint distribution, we sample pairs \( (\hat{\boldsymbol{z}}_1, \hat{\boldsymbol{z}}_2)\) from the encoder’s output for the same input data point, which represents samples from \( p(\hat{\boldsymbol{z}}_1, \hat{\boldsymbol{z}}_2)\). For the marginal distribution, we shuffle \( \hat{\boldsymbol{z}}_2 \) across the batch, generating \( (\hat{\boldsymbol{z}}_1, \hat{\boldsymbol{z}}_2') \), where \( \hat{\boldsymbol{z}}_2' \) is a shuffled version of \( \hat{\boldsymbol{z}}_2 \) from a different data point. This ensures that \( \hat{\boldsymbol{z}}_1 \) and \( \hat{\boldsymbol{z}}_2' \) are independent, approximating the product of the marginals \( p(\hat{\boldsymbol{z}}_1)p(\hat{\boldsymbol{z}}_2) \). Let \( D_\nu \) represent the discriminator parameterized by \( \nu \), trained to distinguish between joint samples \( (\hat{\boldsymbol{z}}_1, \hat{\boldsymbol{z}}_2)\) and marginal samples \( (\hat{\boldsymbol{z}}_1, \hat{\boldsymbol{z}}_2') \). The adversarial loss is defined as

\begin{equation}
\begin{aligned}
\mathcal{L}_{\text{adv}} &= \mathbb{E}_{p(\hat{\boldsymbol{z}}_1, \hat{\boldsymbol{z}}_2)} 
\left[ \log D_\nu(\hat{\boldsymbol{z}}_1, \hat{\boldsymbol{z}}_2) \right] \\
&\quad + \mathbb{E}_{p(\hat{\boldsymbol{z}}_1)p(\hat{\boldsymbol{z}}_2)} 
\left[ \log\left(1 - D_\nu(\hat{\boldsymbol{z}}_1, \hat{\boldsymbol{z}}_2')\right) \right].
\end{aligned}
\end{equation}
This loss encourages \( D_\nu \) to assign high probabilities to true joint samples \( (\hat{\boldsymbol{z}}_1, \hat{\boldsymbol{z}}_2) \) and low probabilities to independent (shuffled) samples \( (\hat{\boldsymbol{z}}_1, \hat{\boldsymbol{z}}_2') \).

To promote disentanglement in the encoder, we add an adversarial penalty to the encoder’s loss. The encoder is trained to fool the discriminator by making the joint distribution \( p(\hat{\boldsymbol{z}}_1, \hat{\boldsymbol{z}}_2) \) indistinguishable from the product of the marginals \( p(\hat{\boldsymbol{z}}_1)p(\hat{\boldsymbol{z}}_2) \). The encoder's loss for disentanglement is defined as

\begin{equation}
   \mathcal{L}_{\text{enc}} = \mathbb{E}_{p(\hat{\boldsymbol{z}}_1, \hat{\boldsymbol{z}}_2)} \left[ \log(1 - D_\nu(\hat{\boldsymbol{z}}_1, \hat{\boldsymbol{z}}_2)) \right]. 
\end{equation}
Minimizing \( \mathcal{L}_{\text{enc}} \) encourages the encoder to make \( \hat{\boldsymbol{z}}_1 \) and \( \hat{\boldsymbol{z}}_2 \) as independent as possible, thereby minimizing the mutual information \( I(\hat{Z}_1; \hat{Z}_2) \). This ensures that the latent representations \( \hat{\boldsymbol{z}}_1\) and \( \hat{\boldsymbol{z}}_2 \) are disentangled, with \( \hat{\boldsymbol{z}}_1 \) capturing task-relevant information and \( \hat{\boldsymbol{z}}_2\) capturing task-irrelevant information.

\begin{algorithm}[]
\caption{Stage 2: Train Task-Relevant Encoder with Contrastive Loss and Discriminator}
\label{alg:train_task_relevant_encoder}
\begin{algorithmic}[1]
\Require $\mathcal{X}_{train}$ (Training dataset), $g_\eta$ (Task-irrelevant encoder), $\lambda$ (Learning rate), $\lambda_{\text{adv}}$ (Discrimnator learning rate),  $\kappa$ (SNR), $\tau$ (Temperature)
\Ensure $f_\theta$
\State Initialize $\theta$, $\psi$, $\nu$
\While{not converged}
    \State $(\boldsymbol{x}, \boldsymbol{y}) \sim \mathcal{X}_{train}$
    \State $\tilde{\boldsymbol{x}}_1 \leftarrow \text{Augment}(\boldsymbol{x})$, $\tilde{\boldsymbol{x}}_2 \leftarrow \text{Augment}(\boldsymbol{x})$
    
    \State $\boldsymbol{z}_1^{(1)} \leftarrow f_\theta(\tilde{\boldsymbol{x}}_1)$,  $\boldsymbol{z}_1^{(2)} \leftarrow f_\theta(\tilde{\boldsymbol{x}}_2)$
    
    \State $\boldsymbol{n}_1 \sim \mathcal{N}(0, \sigma^2 \mathbf{I})$, $\sigma^2 \leftarrow \frac{\mathbb{E}[\|\boldsymbol{z}_1\|^2]}{10^{\frac{\kappa}{10}}}$
    \State $\boldsymbol{n}_2 \sim \mathcal{N}(0, \sigma^2 \mathbf{I})$, $\sigma^2 \leftarrow \frac{\mathbb{E}[\|\boldsymbol{z}_2\|^2]}{10^{\frac{\kappa}{10}}}$

    \State $\hat{\boldsymbol{z}}_1^{(1)} \leftarrow \boldsymbol{z}_1^{(1)} + \boldsymbol{n}_1$, $\hat{\boldsymbol{z}}_1^{(2)} \leftarrow \boldsymbol{z}_1^{(2)} + \boldsymbol{n}_1$
    
    \State $\boldsymbol{h}_1 \leftarrow h_\psi(\hat{\boldsymbol{z}}_1^{(1)})$, $\boldsymbol{h}_2 \leftarrow h_\psi(\hat{\boldsymbol{z}}_1^{(2)})$
    
    \State $\mathcal{L}_{\text{contrast}} \leftarrow \text{Contrastive Loss}(\boldsymbol{h}_1, \boldsymbol{h}_2, \tau)$
    \State $\theta \leftarrow \theta - \lambda \nabla_{\theta} \mathcal{L}_{\text{contrast}}$
    \State $\psi \leftarrow \psi - \lambda \nabla_{\psi} \mathcal{L}_{\text{contrast}}$
    
    \State $\boldsymbol{z}_1 \leftarrow f_\theta(\boldsymbol{x})$, $\boldsymbol{z}_2 \leftarrow g_\eta(\boldsymbol{x})$
    \State $\hat{\boldsymbol{z}}_1 \leftarrow \boldsymbol{z}_1 + \boldsymbol{n}_1$, $\hat{\boldsymbol{z}}_2 \leftarrow \boldsymbol{z}_2 + \boldsymbol{n}_2$
    
    \State $\hat{\boldsymbol{z}}_2'$ $\leftarrow$ Shuffle $\hat{\boldsymbol{z}}_2$ across the batch
    \State $\mathcal{L}_{\text{adv}} \leftarrow \log D_\nu(\hat{\boldsymbol{z}}_1, \hat{\boldsymbol{z}}_2) + \log(1 - D_\nu(\hat{\boldsymbol{z}}_1, \hat{\boldsymbol{z}}_2'))$
    \State $\nu \leftarrow \nu - \lambda_{\text{adv}} \nabla_{\nu} \mathcal{L}_{\text{adv}}$
    
    \State $\mathcal{L}_{\text{enc}} \leftarrow  \log(1 - D_\nu(\hat{\boldsymbol{z}}_1, \hat{\boldsymbol{z}}_2)) $
    \State $\theta \leftarrow \theta - \lambda \nabla_{\theta} \mathcal{L}_{\text{enc}}$
\EndWhile
\State Discard $h_\psi$, $D_\nu$ $g_\eta$
\State Freeze $\theta$
\end{algorithmic}
\end{algorithm}

\subsection{Classification Task}

The final downstream task is classification, where the goal is to predict the label \( Y \) from the encoded features \( \hat{Z}_1 \). We use a simple feed-forward neural network classifier \( q_\phi(\boldsymbol{y}|\hat{\boldsymbol{z}}_1) \), parameterized by \( \phi \), and trained with a cross-entropy loss. The classifier takes as input the task-relevant features \( \hat{\boldsymbol{z}}_1 \) and is optimized to minimize the following cross-entropy loss:

\begin{equation}
    \mathcal{L}_{\text{class}} = - \mathbb{E}_{p(\boldsymbol{x}, \boldsymbol{y})} \left[ \sum_{c=1}^{C} y_c \log q_\phi(y_c | \hat{\boldsymbol{z}}_1) \right],
\end{equation}
where \( C \) is the number of classes, and \( y_c \) is the ground truth one-hot encoded label for class \( c \).

\begin{algorithm}[]
\caption{Stage 3: Train Classifier on Frozen Task-Relevant Encoder}
\label{alg:train_classifier}
\begin{algorithmic}[1]
\Require$\mathcal{X}_{train}$ (Training dataset), $f_\theta$ (Task relevant encoder), learning rate $\lambda$, $\kappa$ (SNR)
\Ensure Trained classifier $q_\phi$
\State Initialize $\phi$
\While{not converged}
    \State Sample $(\boldsymbol{x}, \boldsymbol{y}) \sim \mathcal{X}_{train}$
    \State $\boldsymbol{z}_1 \leftarrow f_\theta(\boldsymbol{x})$ 
    \State $\boldsymbol{n} \sim \mathcal{N}(0, \sigma^2 \mathbf{I})$, $\sigma^2 \leftarrow \frac{\mathbb{E}[\|\boldsymbol{z}_1\|^2]}{10^{\frac{\kappa}{10}}}$
  \State $\hat{\boldsymbol{z}}_1 \leftarrow \boldsymbol{z}_1 + \boldsymbol{n}$
    \State $\hat{\boldsymbol{y}} \leftarrow q_\phi(\hat{\boldsymbol{z}}_1)$
    \State $\mathcal{L}_{\text{class}} \leftarrow - \sum_{i=1}^{C} \boldsymbol{y}_i \log \hat{\boldsymbol{y}}_i$
    \State $\phi \leftarrow \phi - \lambda  \nabla_{\phi} \mathcal{L}_{\text{class}}$
\EndWhile
\end{algorithmic}
\end{algorithm}

\subsection{Training Procedure}

Training a complex system with many different components and loss function must be performed carefully to ensure that each stage achieves its goal without interfering with other objectives and that the gradients flow appropriately. The training is done in multiple stages, each targeting a different part of the system. Below, we outline the step-by-step procedure used to train our model and the associated algorithms. 

    \noindent \textit{Stage 1: Training the Task-Irrelevant Encoder:} In the first stage, we train the task-irrelevant encoder \( g_\eta \) by pairing it with a reconstructor \( r_\omega \). The reconstructor \( r_\omega \), learns to reconstruct an image by using the encoded representations  from \( g_\eta \) as well as the label information$\boldsymbol{y}$.  This encourages \( g_\eta \) to focus on capturing the parts of the input that are not necessary for the downstream classification task by minimizing the reconstruction loss. The reconstructor is discarded, and the parameters of \( g_\eta \) are frozen after training to preserve the task-irrelevant features for later use. This procedure is outlined in Algorithm~\ref{alg:train_encoder_reconstructor}.
    
    \noindent \textit{Stage 2: Training the Task-Relevant Encoder with Contrastive Loss and Discriminator:} After freezing \( g_\eta \), the task-relevant encoder \( f_\theta \) is trained in this stage. We use both a contrastive loss to ensure that \( f_\theta \) captures class-discriminative features and an adversarial loss to enforce disentanglement between the task-relevant encoder \( f_\theta \) feature vector and the task-irrelevant encoder \( g_\eta \) feature vector. The task-relevant encoder is trained using augmented views of the input for contrastive learning and through adversarial training with the discriminator \( D_\nu \). The training process alternates between updating the contrastive loss and updating the discriminator and encoder to ensure disentanglement. The details of this stage are described in Algorithm~\ref{alg:train_task_relevant_encoder}.
    
    \noindent \textit{Stage 3: Training the Classifier on the Frozen Task-Relevant Encoder:} Once disentanglement is achieved, we discard the projection head \( h_\psi \), the discriminator \( D_\nu \), and the task-irrelevant encoder \( g_\eta \), leaving only the frozen task-relevant encoder \( f_\theta \). In this final stage, we train the classifier \( q_\phi \) on top of \( f_\theta \) for the downstream classification task. The classifier is trained with the cross-entropy loss, ensuring that it can utilize the task-relevant features \( f_\theta \) for accurate classification. The classifier training procedure is outlined in Algorithm~\ref{alg:train_classifier}.
    
The separation of training stages is for stability, and also reflects a modular decomposition of the CLAD objective. Each stage isolates the gradient flow to the component most relevant for that term, making the multi-stage training strategy a practical approximation to optimizing \( \mathcal{L}_{\text{CLAD}} \) holistically.

\subsection{Information Retention Index Across Different Methods}

To assess how much information \( \hat{Z} \) retains about input \( X \), we estimate \( I(\hat{Z}; X) \), which quantifies the informativeness of the latent representation \( \hat{Z} \) for reconstructing the original input \( X \). Since direct computation of mutual information is intractable, we adopt a reconstruction-based proxy \cite{hjelmlearning} to compute the IRI across different methods.

Assume that the reconstruction loss \( \mathcal{L}_{\text{recon}}(\boldsymbol{x}|\hat{\boldsymbol{z}}) \), parameterized by a reconstructor \( r_\gamma(\cdot) \) with parameters \( \gamma \), denotes the expected error for reconstructing \( \boldsymbol{x} \) from the latent representation \( \hat{\boldsymbol{z}} \). The mutual information \( I(\hat{Z}; X) \) can be bounded as follows:
\begin{equation}
   I(\hat{Z}; X) = H(X) - H(X | \hat{Z}) \geq H(X) - \mathbb{E}_{p(\boldsymbol{x}, \hat{\boldsymbol{z}})} \left[ \mathcal{L}_{\text{recon}}(\boldsymbol{x}|\hat{\boldsymbol{z}}) \right], 
\end{equation}
where \( H(X) \) represents the entropy of the input, and \( \mathcal{L}_{\text{recon}}(\boldsymbol{x}|\hat{\boldsymbol{z}}) \) is the reconstruction loss. Therefore, one can compute \( I(\hat{Z}; X) \) by minimizing the reconstruction error as follows:
\begin{equation}
 I(\hat{Z}; X) \geq H(X) - \min_{\gamma} \mathcal{L}_{\text{recon}}^\gamma(\boldsymbol{x}|\hat{\boldsymbol{z}}).   
\end{equation}

In practice, for each task-oriented communication method we evaluate, the corresponding encoder parameters are frozen, and a reconstructor \( r_\gamma \) is trained to minimize the reconstruction loss \({L}_{\text{recon}}(\boldsymbol{x}|\hat{\boldsymbol{z}})\). The reconstructor is trained using mean squared error (MSE) as the loss function, and we evaluate the quality of the reconstructions using the structural similarity index measure (SSIM) \cite{wang2004image}. SSIM serves as a proxy for the total mutual information between the input and the representation and can indicate the
amount of encoded pixel-level information. It has been shown empirically that SSIM correlated with mutual information \cite{hjelmlearning}. We drop $H(X)$ from our calculations as it is a constant. 

Unlike MSE, which only measures pixel-wise differences, SSIM accounts for luminance, contrast, and structural information, providing a better perceptual measure of image quality. This makes SSIM a suitable proxy to indicate how much useful information from \( X \) is retained in \( \hat{Z} \). The SSIM between two images \( x \) and \( \hat{x} \) is given by

\begin{equation}
  \text{SSIM}(x, \hat{x}) = \frac{(2\mu_x \mu_{\hat{x}} + c_1)(2\sigma_{x\hat{x}} + c_2)}{(\mu_x^2 + \mu_{\hat{x}}^2 + c_1)(\sigma_x^2 + \sigma_{\hat{x}}^2 + c_2)},  
\end{equation}
where \( \mu_x \) and \( \mu_{\hat{x}} \) are the mean intensities of the original and reconstructed images, \( \sigma_x^2 \) and \( \sigma_{\hat{x}}^2 \) are their variances, and \( \sigma_{x\hat{x}} \) is the covariance between them. The constants \( c_1 \) and \( c_2 \) stabilize the division to avoid near-zero values. 

The SSIM has a range from  -1 and 1, with values closer to 1 indicating higher structural similarity. By focusing on perceptual quality rather than pixel-wise differences, we find that SSIM provides a more accurate measure of the retained information in the latent representation. Formally, we define the IRI for a given task-oriented communication system \(i\) as

\begin{equation}
    \text{IRI}_i = \text{SSIM}(\boldsymbol{x}, r_{\gamma_i}(\hat{\boldsymbol{z}})),
\end{equation}
where \( r_{\gamma_i} \) refers to the reconstructor specifically trained for system \(i\).

To compare the different systems fairly, we ensure the following conditions:
\begin{itemize}
    \item The same decoder architecture is used for each system, ensuring consistency across the experiments.
    \item All reconstructors are trained with the same settings and hyperparameters for the same number of epochs.
    \item We train all reconstructors on the same training set and to ensure a valid comparison, we assess the reconstruction performance on the same testing set.
\end{itemize}
By comparing the IRI scores on the reconstructed images, we can capture the information retention across different methods. The higher the IRI, the more information \( Z \) retains about \( X \), allowing us to quantify the informativeness and minimality of the learned representations. Algorithm \ref{alg:estimate_I_ZX_SSIM} provides a detailed procedure to compute the IRI by leveraging the correlation between reconstructed and original inputs to approximate informativeness and minimality.

\begin{figure}[t]
    \centering
    \includegraphics[width=0.9\linewidth]{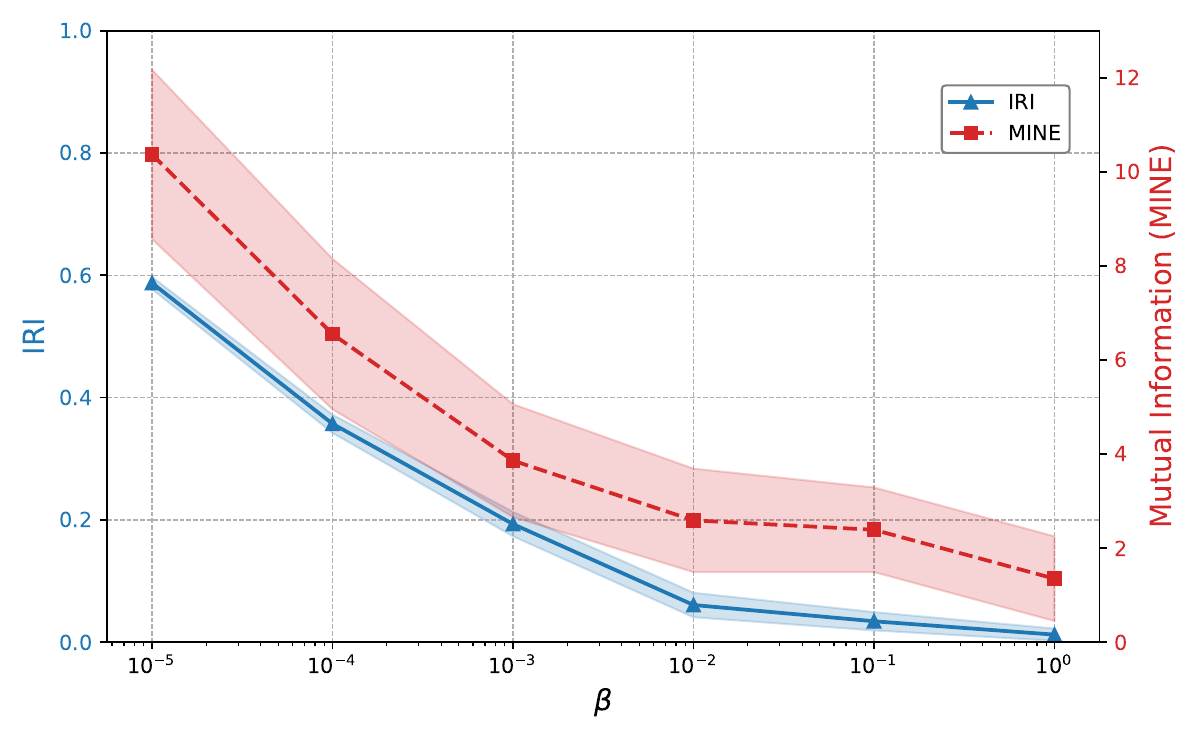}
    \caption{Comparison of IRI and MINE estimates across different values of \( \beta \) in the VIB objective. Both metrics exhibit similar trends in retained information, but MINE shows significantly higher variance. Shaded regions represent the standard deviation over multiple runs.}
    \label{fig:iri_mine_comparison}
\end{figure}

Although Mutual Information Neural Estimation (MINE) is a widely-used method for estimating the mutual information \cite{belghazi2018mine}, it suffers from several well-documented drawbacks that limit its effectiveness for benchmarking in semantic communication systems \cite{song2019understanding}. MINE is prone to high variance, requires careful tuning of many different hyperparameters, and scales poorly in high-dimensional latent spaces all of which can lead to unstable or misleading estimates. These issues are especially problematic when comparing different models or compression levels under consistent settings. In contrast, IRI allows for consistent, fair benchmarking with minimal assumptions and training overhead. We view IRI as a relative measure, similar to a diagnostic tool, that enables empirical analysis of minimality and informativeness in task-oriented systems.  As illustrated in Figure~\ref{fig:iri_mine_comparison}, IRI follows the same general trend as MINE but with substantially reduced variance, providing a more stable and reliable metric for evaluating privacy-relevant information retention in 6G-IoT task-oriented communication systems.

 We note that while SSIM is effective in image-based applications, it is inherently limited to image data. For non-visual modalities such as text or tabular data, alternative reconstruction-based metrics (e.g., BLEU score) can be adopted to extend the IRI framework to broader task domains.

\begin{algorithm}[]
\caption{Computing IRI}
\label{alg:estimate_I_ZX_SSIM}
\begin{algorithmic}[1]
\Require $\mathcal{X}_{train}$, $\mathcal{X}_{test}$ frozen encoders $\{f_{\theta_i}\}_{i=1}^M$, learning rate $\lambda$, $\kappa$ (SNR)
\Ensure IRI for each system $i$
\State Initialize reconstructors $\{r_{\gamma_i}\}_{i=1}^M$
\For{$i = 1$ to $M$}
    \State Freeze $\theta_i$
    \While{not converged}
        \State Sample $(\boldsymbol{x}, \boldsymbol{y}) \sim \mathcal{X}_{train}$
        \State $\boldsymbol{z} \leftarrow f_{\theta_i}(\boldsymbol{x})$
            \State $\boldsymbol{n} \sim \mathcal{N}(0, \sigma^2 \mathbf{I})$, $\sigma^2 \leftarrow \frac{\mathbb{E}[\|\boldsymbol{z}_1\|^2]}{10^{\frac{\kappa}{10}}}$
              \State $\hat{\boldsymbol{z}}_1 \leftarrow \boldsymbol{z}_1 + \boldsymbol{n}$
        \State $\hat{\boldsymbol{x}} \leftarrow r_{\gamma_i}(\hat{\boldsymbol{z}_1})$
        \State $\mathcal{L}_{\text{recon}} \leftarrow \| \boldsymbol{x} - \hat{\boldsymbol{x}} \|^2$
        \State $\gamma_i \leftarrow \gamma_i - \lambda_{\text{rec}} \nabla_{\gamma_i} \mathcal{L}_{\text{recon}}$
    \EndWhile
\EndFor
\For{$i = 1$ to $M$}
    \State Sample $(\boldsymbol{x}, \boldsymbol{y}) \sim \mathcal{X}_{test}$
            \State $\boldsymbol{z} \leftarrow f_{\theta_i}(\boldsymbol{x})$
            \State $\boldsymbol{n} \sim \mathcal{N}(0, \sigma^2 \mathbf{I})$, $\sigma^2 \leftarrow \frac{\mathbb{E}[\|\boldsymbol{z}_1\|^2]}{10^{\frac{\kappa}{10}}}$
              \State $\hat{\boldsymbol{z}}_1 \leftarrow \boldsymbol{z}_1 + \boldsymbol{n}$
        \State $\hat{\boldsymbol{x}} \leftarrow r_{\gamma_i}(\hat{\boldsymbol{z}_1})$
\State $\text{IRI}_i \gets \text{SSIM}(\boldsymbol{x}, \hat{\boldsymbol{x}})$

\EndFor
\end{algorithmic}
\end{algorithm}

\section{Experimental Evaluations and Discussion}
In this section, we present the experimental setup used to evaluate CLAD. We use image classification as a representative task to illustrate the core concept of the proposed methods, developing an end-to-end learning framework that extracts low-dimensional, task-relevant, privacy-preserving, and channel-robust latent representations for trustworthy 6G-IoT applications. Importantly, the proposed framework is not limited to image classification alone. We start by describing the datasets used in our experiments, followed by a discussion of the baseline methods, neural architectures, and the experimental setup. Finally, we present detailed evaluations and analysis of the results.\footnote{The source code, models and results are available at https://github.com/OmarErak/CLAD}.

\subsection{Experimental Setup}

\subsubsection{Datasets}
The Colored MNIST and Colored FashionMNIST datasets are extensions of the standard MNIST \cite{6296535} and FashionMNIST \cite{xiao2017fashionmnistnovelimagedataset} datasets, each consisting of 60,000 28x28 grayscale images. In Colored MNIST, handwritten digits (0-9) are overlaid on colored backgrounds, while in Colored FashionMNIST, clothing items from 10 categories (e.g., T-shirts, coats, shoes) are similarly displayed on colored backgrounds. The introduction of background colors adds additional task-irrelevant information, creating a more challenging setup for the model to disentangle task-relevant features relevant to classifying digits or clothing items from background-related attributes. Furthermore, incorporating background color labels enables the evaluation of attribute inference attacks, where an adversary is trained to predict background color. This setup provides insight into the model's ability to protect against such attacks while maintaining disentanglement between task-relevant and task-irrelevant features. To further evaluate the scalability and robustness of the proposed method on more complex and natural datasets, we also incorporate CIFAR-10 \cite{krizhevsky2009learning}. CIFAR-10 consists of 60,000 32x32 color images in 10 classes, including airplanes, cars, birds, cats, and other natural objects. Compared to MNIST-based datasets, CIFAR-10 introduces significantly more visual variability and semantic richness, making it a more challenging benchmark.

\subsubsection{Neural Network Architectures}
To simulate realistic 6G-IoT scenarios, we adopt deep neural network (DNN) architectures for both the task-relevant encoder at the transmitter and the downstream classifier at the receiver. These networks consist of convolutional and fully connected layers, structured around a latent dimension \( d \). The encoder and classifier architectures, outlined in Table~\ref{colored_mnist_classification_structure} and Table~\ref{cifar10_encoder_classifier_structure}, are used consistently across all evaluated methods to ensure fair and reproducible comparisons. To compute the IRI, we additionally employ reconstructor networks that attempt to recover the original input \( \boldsymbol{x} \) from the latent representation \( \hat{\boldsymbol{z}} \). The architectures for these reconstructors are provided in Table~\ref{reconstructor_structure} and Table~\ref{cifar10_reconstructor_structure}, and are applied uniformly across all methods during IRI evaluation.

\subsubsection{Channel Conditions} 
We evaluate the performance of CLAD compared to baseline methods using an AWGN channel model due to its widespread adoption. Specifically, we consider training and testing the models at identical SNRs, ranging from -6 dB to 12 dB. This setting allows us to assess the robustness of each method across different noise levels in a controlled manner. For each SNR value, we train the models over multiple runs and average the results to mitigate any randomness introduced during training.  In the following experiments, we simulate a constrained wireless edge scenario by setting the channel bandwidth to 12.5kHz and the symbol rate to 9,600 baud, reflecting practical limitations in edge communication environments.

\subsection{Baselines}
In our experiments, we compare the proposed method CLAD against three baselines: DeepJSCC \cite{bourtsoulatze2019deep}, VIB and\cite{alemi2017deep, shao2021learning} and  information bottleneck and adversarial learning (IBAL) \cite{wang2024privacy}. These methods provide a benchmark for task-oriented communication systems, helping to evaluate the effectiveness of our approach in terms of privacy and downstream task performance.

\subsubsection{DeepJSCC}
DeepJSCC  is a neural network-based approach that optimizes the encoding of data for transmission over noisy channels. For our task-oriented scenario, DeepJSCC is trained with cross-entropy loss for classification rather than reconstruction, and it does not explicitly discard task-irrelevant information. As a result, it serves as a baseline for how well the encoded representation performs without feature disentanglement.

\subsubsection{Variational Information Bottleneck (VIB)}
The VIB framework aims to compress the input \( X \) into  latent representation \( Z \) while retaining sufficient information for predicting \( Y \). VIB balances \( I(Z; X) \) and \( I(Z; Y) \) via hyperparameter \( \beta \). In our experiments, we provide results for different values of \( \beta \) to illustrate how varying the trade-off between compression and task relevance impacts task performance, IRI, and privacy. The Variational Feature Encoding (VFE) method in  \cite{shao2021learning} is based on VIB, and therefore the VIB results presented are synonymous to VFE.

\subsubsection{Information Bottleneck and Adversarial Learning (IBAL)}
IBAL is a task-oriented semantic communication approach that modifies the traditional VIB objective by incorporating an additional distortion constraint. Specifically, IBAL optimizes a composite loss function that balances the original variational information bottleneck objective with an MSE term that is maximized, which encourages poor reconstructions and thereby enhances resistance to model inversion attacks.

By benchmarking against these three baselines, we show that CLAD more effectively extracts task-relevant features, suppresses task-irrelevant information, enhances downstream classification performance, and offers stronger privacy guarantees.

\begin{table}[htb]\scriptsize
\caption{DNN Structure for the transmitter (encoder) and the receiver (classifier) used for Colored MNIST and FashionMNIST }
\label{colored_mnist_classification_structure}
\centering
\renewcommand{\arraystretch}{1.75}
\begin{tabular}{c|c|c}
\hline
&  \textbf{Layer}                                                          & \begin{tabular}[c]{@{}c@{}}\textbf{Output}\\\textbf{dimensions}\end{tabular}\\
\hline
\textbf{Transmitter}                                              & \begin{tabular}[c]{@{}c@{}}Conv Layer+ReLU\\MaxPool Layer\\Conv Layer+ReLU\\MaxPool Layer\\Fully Connected (Flatten)\end{tabular}                                                     & \begin{tabular}[c]{@{}c@{}}32$\times$28$\times$28\\32$\times$14$\times$14\\64$\times$14$\times$14\\64$\times$7$\times$7\\$d$\end{tabular} \\
\hline
\begin{tabular}[c]{@{}c@{}}\textbf{Receiver}\end{tabular} & \begin{tabular}[c]{@{}c@{}}Fully Connected (FC)\\Fully Connected (FC)\\Fully Connected + Softmax\end{tabular} & \begin{tabular}[c]{@{}c@{}}512\\256\\10\end{tabular}  \\
\hline
\end{tabular}
\label{arch}
\end{table}

\begin{table}[htb]\scriptsize
\caption{Architecture settings for the reconstructors used with Colored MNIST and FashionMNIST to evaluate IRI}
\label{reconstructor_structure}
\centering
\renewcommand{\arraystretch}{1.75}
\begin{tabular}{c|c|c}
\hline
&  \textbf{Layer name}                                                          & \begin{tabular}[c]{@{}c@{}}\textbf{Output}\\\textbf{dimensions}\end{tabular}\\
\hline
\multirow{5}{*}{\textbf{Reconstructor}} 
    & Fully Connected (FC)                & 128$\times$7$\times$7      \\ \cline{2-3} 
    & Deconv Layer + ReLU                 & 64$\times$14$\times$14     \\ \cline{2-3} 
    & Deconv Layer + ReLU                 & 32$\times$28$\times$28     \\ \cline{2-3} 
    & Deconv Layer + ReLU                 & 16$\times$28$\times$28     \\ \cline{2-3} 
    & Deconv Layer + Sigmoid              & 3$\times$28$\times$28      \\ 
\hline
\end{tabular}
\end{table}

\begin{table}[htb]\scriptsize
\caption{DNN structure for the transmitter (encoder) and the receiver (classifier) used for CIFAR-10}
\label{cifar10_encoder_classifier_structure}
\centering
\renewcommand{\arraystretch}{1.5}
\begin{tabular}{c|c|c}
\hline
& \textbf{Layer} & \begin{tabular}[c]{@{}c@{}}\textbf{Output} \\ \textbf{Dimensions}\end{tabular} \\
\hline
\textbf{Transmitter} & \begin{tabular}[c]{@{}c@{}}Conv + ReLU $\times$2\\ResNet Block\\Conv + ReLU $\times$2\\Reshape + FC + Tanh\end{tabular} 
& \begin{tabular}[c]{@{}c@{}}128$\times$32$\times$32\\128$\times$16$\times$16\\4$\times$4$\times$4\\$d$\end{tabular} \\
\hline
\textbf{Receiver} & \begin{tabular}[c]{@{}c@{}}FC + ReLU + Reshape\\Conv + ReLU $\times$2\\ResNet Block\\Pooling Layer\\FC + Softmax\end{tabular} 
& \begin{tabular}[c]{@{}c@{}}64$\times$4$\times$4\\512$\times$4$\times$4\\512$\times$4$\times$4\\512\\10\end{tabular} \\
\hline
\end{tabular}
\end{table}

\begin{table}[htb]\scriptsize
\caption{Architecture settings for the reconstructors used with CIFAR-10 to evaluate IRI}
\label{cifar10_reconstructor_structure}
\centering
\renewcommand{\arraystretch}{1.75}
\begin{tabular}{c|c|c}
\hline
& \textbf{Layer name} & \begin{tabular}[c]{@{}c@{}}\textbf{Output}\\\textbf{dimensions}\end{tabular} \\
\hline
\multirow{6}{*}{\textbf{Reconstructor}} 
    & Fully Connected (FC) + ReLU         & 512$\times$4$\times$4      \\ \cline{2-3}
    & Deconv (512$\rightarrow$256) + ReLU & 256$\times$8$\times$8      \\ \cline{2-3}
    & Deconv (256$\rightarrow$128) + ReLU & 128$\times$16$\times$16    \\ \cline{2-3}
    & Deconv (128$\rightarrow$64) + ReLU  & 64$\times$32$\times$32     \\ \cline{2-3}
    & Conv (64$\rightarrow$3) + Sigmoid   & 3$\times$32$\times$32      \\ 
\hline
\end{tabular}
\end{table}

\subsection{Evaluation Metrics}
To assess the effectiveness of CLAD, we employ four key evaluation metrics: classification accuracy for task performance, IRI, attribute inference attack accuracy, and model inversion attacks for privacy assessment. Furthermore, all methods are evaluated across a range of channel SNRs to examine their robustness under dynamic transmission conditions.

\subsubsection{Task Performance (Accuracy)}
The primary evaluation metric for task performance is classification accuracy. It measures the ability of classifier to predict the label \( \boldsymbol{y} \) from \( \boldsymbol{\hat{z}} \). Accuracy is calculated as the ratio of correctly classified instances to the total number of instances:

\begin{equation}
    \text{Accuracy} =  \frac{1}{N} \sum_{i=1}^{N} \mathds{1}(\hat{y}_i = y_i),
\end{equation}
where \( N \) is the total number of samples, \( \hat{y}_i \) is the predicted label, and \( y_i \) is the ground truth label.

\subsubsection{Information Retention Index (IRI)}
To quantify the amount of information retained in the encoded representation \( \hat{Z} \), we use our proposed method to compute the IRI. This measures how much information from the input \( X \) is present in the encoded representation \( \hat{Z} \), which helps assess the compression of the representation. By comparing IRI across different methods, we can evaluate how effectively each method discards task-irrelevant information.

\begin{figure*}[]
    \centering
    \subfloat[]{
        \centering
        \includegraphics[width=0.31\linewidth]{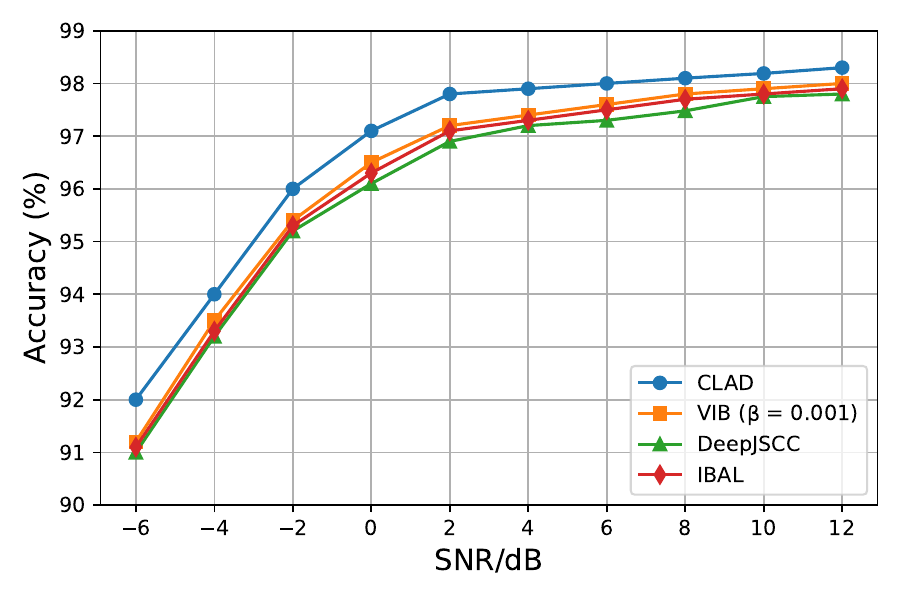}
        \label{subfig:MNIST_SNR}
    }
    \subfloat[]{
        \centering
        \includegraphics[width=0.31\linewidth]{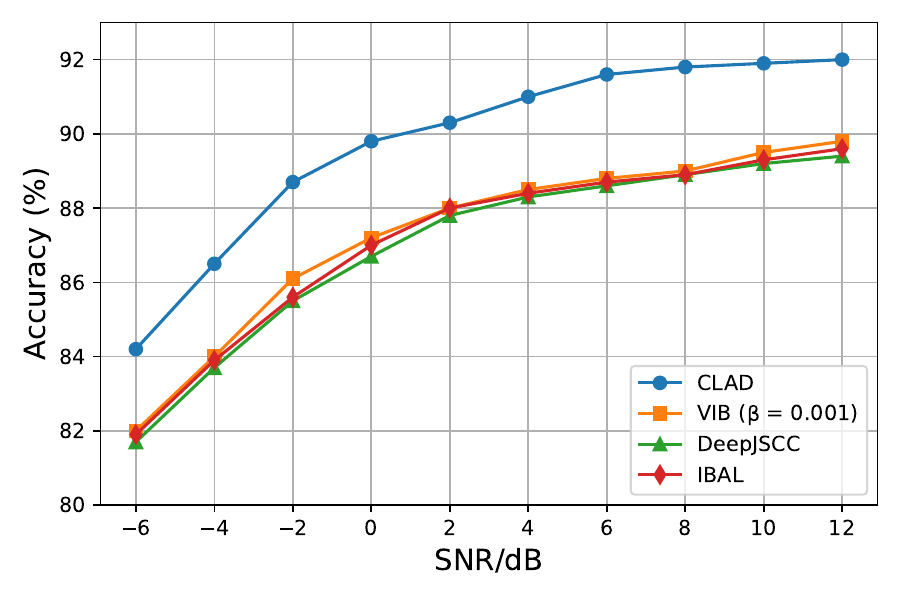}
        \label{subfig:FMNIST_SNR}
    }
    \subfloat[]{
        \centering
        \includegraphics[width=0.31\linewidth]{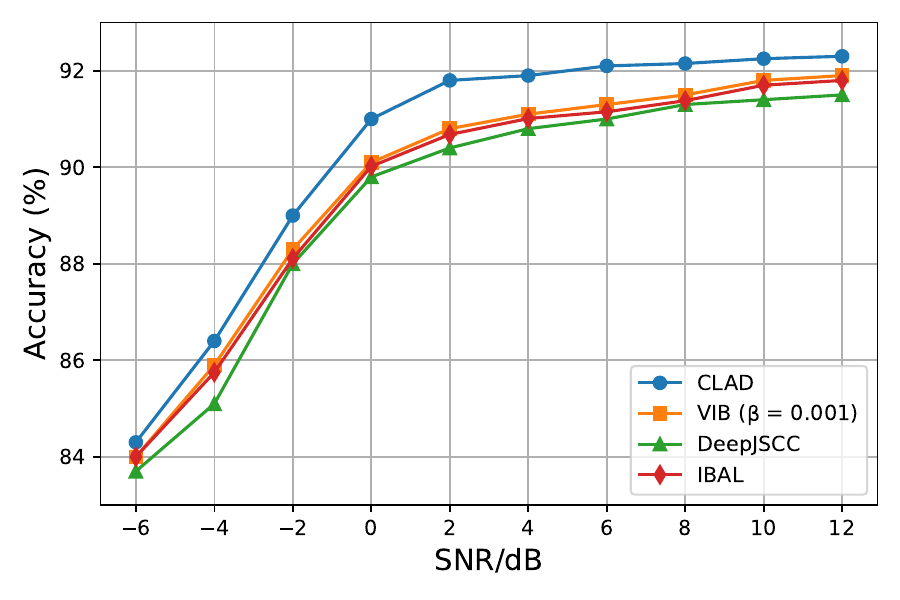}
        \label{subfig:CIFAR10_SNR}
    }
    \caption{Accuracy at different SNRs for (a) the Colored MNIST dataset, (b) the Colored FashionMNIST dataset and (c) the CIFAR-10 dataset .}
    \label{fig:graph_results_1}
\end{figure*}

\begin{figure*}[]
    \centering
    \subfloat[]{
        \centering
        \includegraphics[width=0.31\linewidth]{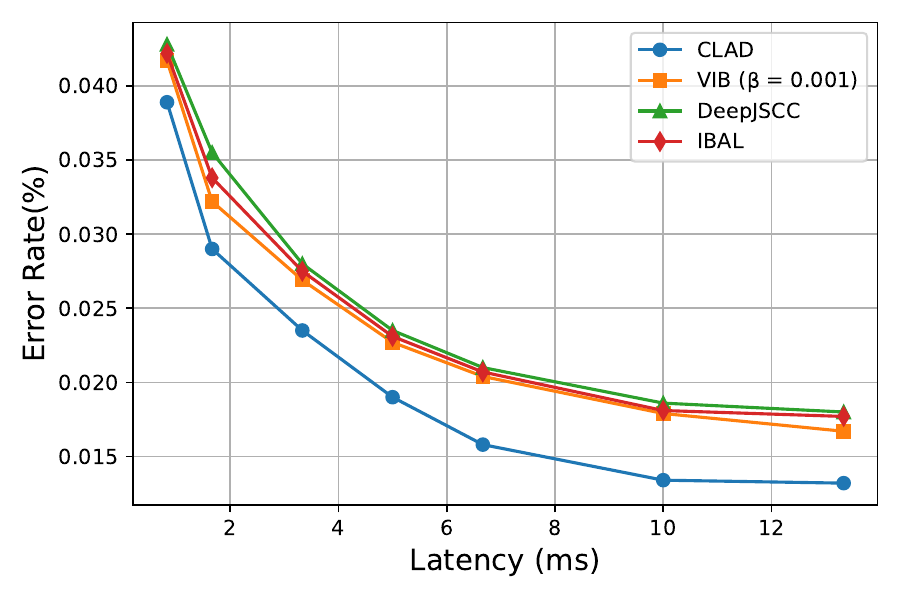}
        \label{subfig:MNIST_LAT}
    }
    \subfloat[]{
        \centering
        \includegraphics[width=0.31\linewidth]{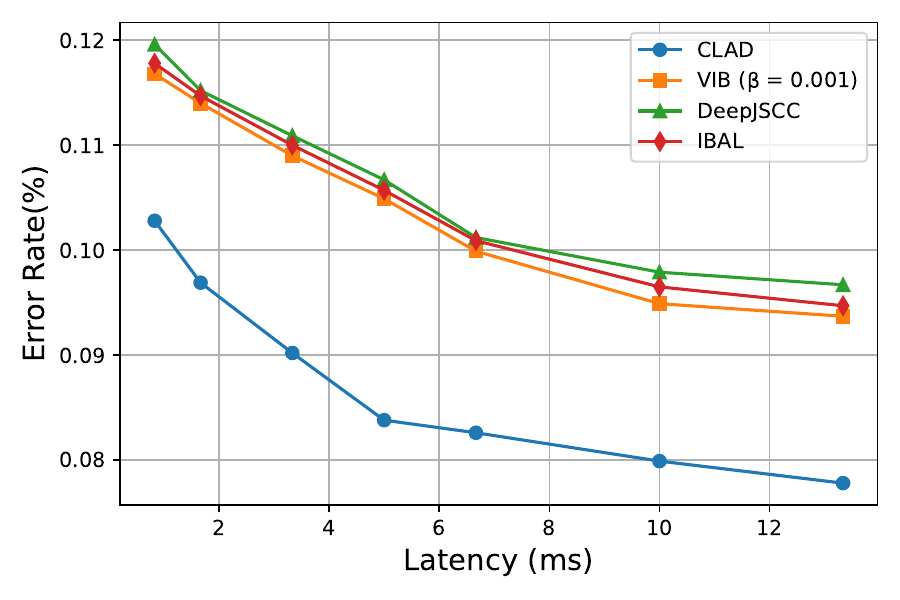}
        \label{subfig:FMNIST_LAT}
    }
    \subfloat[]{
        \centering
        \includegraphics[width=0.31\linewidth]{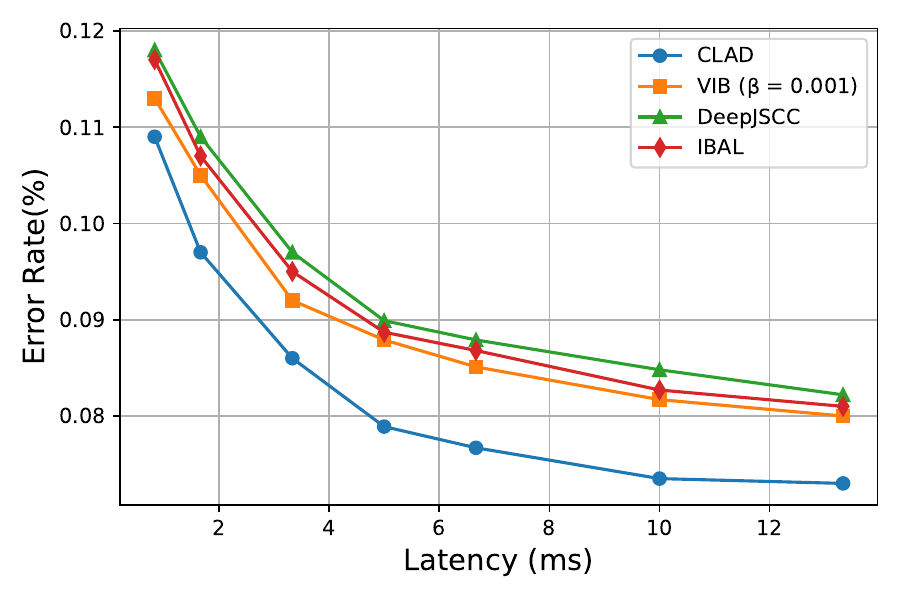}
        \label{subfig:CIFAR10_LAT}
    }
    \caption{Rate-distortion curves for (a) the Colored MNIST dataset, (b) the Colored FashionMNIST dataset and (c) the CIFAR-10 dataset. SNR is set to 12dB.}
    \label{fig:graph_results_2}
\end{figure*}

\subsubsection{Attribute Inference Attack}
In addition to task performance and information retention, we also evaluate privacy by comparing the vulnerability of different methods to attribute inference attacks. An attribute inference attack aims to recover sensitive or irrelevant information about the input, such as background color, from the encoded representation \( \hat{\boldsymbol{z}} \).

Given the encoded representation \( \hat{\boldsymbol{z}} \), the adversary seeks to predict the background color of the image.  If \( \hat{\boldsymbol{z}}  \) contains significant task-irrelevant information, the adversary will be able to classify the background color with high accuracy. To evaluate this, we train a background color classifier (with the same architecture as the task relevant classifier in Table \ref{arch}) on the encoded representation \( \hat{\boldsymbol{z}}  \) and report its classification accuracy. A higher accuracy in the attribute inference attack implies more task-irrelevant information is retained in \( \hat{\boldsymbol{z}} \), indicating weaker privacy guarantees.

\subsubsection{Model Inversion Attack}
To further evaluate privacy leakage, we consider model inversion attacks, which aim to reconstruct the original input data \( \boldsymbol{x} \) from the encoded representation \( \hat{\boldsymbol{z}} \). This type of attack simulates an adversary that gains access to the transmitted latent representation and trains a decoder to recover the input image. In our setup, we follow a black-box attack scenario, where the adversary does not have access to the encoder or its training data. Specifically, we split the dataset such that 4/5 of the data is used to train and test the encoder and classifier models, while the remaining 1/5 is reserved for training the adversary. The adversary learns a mapping from \( \hat{\boldsymbol{z}} \) to the corresponding input image using only this held-out subset. A visually accurate reconstruction indicates that \( \hat{\boldsymbol{z}} \) still encodes significant low-level input features, implying weaker privacy preservation. Similar to \cite{wang2024privacy}, we use SSIM as a performance measure for the model inversion attacks.

\begin{table}[htb]\scriptsize
\caption{Evaluation of different methods on Colored MNIST dataset at SNR = 12 dB, under a latency constraint of $t \leq 6.67$ ms.}
\label{tab:colored_mnist_evaluation}
\centering
\renewcommand{\arraystretch}{2}
\begin{tabular}{c|c|c|c}
\hline
\textbf{Method} & \begin{tabular}[c]{@{}c@{}}\textbf{Classification}\\\textbf{Accuracy (\%)}\end{tabular} & \begin{tabular}[c]{@{}c@{}}\textbf{IRI} \\ \end{tabular} & \begin{tabular}[c]{@{}c@{}}\textbf{Adversarial}\\\textbf{Accuracy (\%)}\end{tabular} \\ \hline
\textbf{DeepJSCC} & 97.96 & 0.608 & 79.16 \\ 
\hline
\textbf{VIB (Beta=0.0001)} & 98.01 & 0.3931 & 52.12 \\
\hline
\textbf{VIB (Beta=0.001)} & 97.90 & 0.1931 & 34.09 \\
\hline
\textbf{VIB (Beta=0.01)} & 96.93 & 0.0608 & 22.96 \\ 
\hline
\textbf{VIB (Beta=0.1)} & 93.29 & 0.0342 & 17.56 \\
\hline
\textbf{VIB (Beta=1)} & 11.36 & 0.0123 & 13.52 \\
\hline
\textbf{IBAL} & 97.63 & 0.1762 & 29.86 \\
\hline
\textbf{CLAD (Ours)} & \textbf{98.42} & \textbf{0.039} & \textbf{19.83} \\ 
\hline
\end{tabular}
\label{MNIST_RES}
\end{table}

\begin{table}[htb]\scriptsize
\caption{Evaluation of different methods on Colored FashionMNIST dataset at SNR = 12 dB, under a latency constraint of $t \leq 6.67$ ms.}
\label{tab:colored_fmnist_evaluation}
\centering
\renewcommand{\arraystretch}{2}
\begin{tabular}{c|c|c|c}
\hline
\textbf{Method} & \begin{tabular}[c]{@{}c@{}}\textbf{Classification}\\\textbf{Accuracy (\%)}\end{tabular} & \begin{tabular}[c]{@{}c@{}}\textbf{IRI} \\\end{tabular} & \begin{tabular}[c]{@{}c@{}}\textbf{Adversarial}\\\textbf{Accuracy (\%)}\end{tabular} \\ \hline
\textbf{DeepJSCC} & 89.28 & 0.5958 & 79.52 \\ 
\hline
\textbf{VIB (Beta=0.0001)} & 90.02 & 0.3172 & 58.03 \\
\hline
\textbf{VIB (Beta=0.001)} & 89.30 & 0.2562 & 47.00 \\
\hline
\textbf{VIB (Beta=0.01)} & 86.98 & 0.0707 & 23.55 \\ 
\hline
\textbf{VIB (Beta=0.1)} & 81.82 & 0.0497 & 17.06 \\
\hline
\textbf{VIB (Beta=1)} & 11.28 & 0.0101 & 12.82 \\
\hline
\textbf{IBAL} & 89.15 & 0.1842 & 32.06 \\
\hline
\textbf{CLAD (Ours)} & \textbf{91.74} & \textbf{0.0587} & \textbf{19.33} \\ 
\hline
\end{tabular}
\label{FMNIST_Res}
\end{table}

\begin{table}[htb]\scriptsize
\caption{Evaluation of different methods on CIFAR10 dataset at SNR = 12 dB, under a latency constraint of $t \leq 6.67$ ms.}
\label{tab:cifar10_evaluation}
\centering
\renewcommand{\arraystretch}{2}
\begin{tabular}{c|c|c}
\hline
\textbf{Method} & \begin{tabular}[c]{@{}c@{}}\textbf{Classification} \\ \textbf{Accuracy (\%)} \end{tabular} & \textbf{IRI} \\
\hline
\textbf{DeepJSCC} & 91.21 & 0.2043 \\ 
\hline
\textbf{VIB (Beta=0.0001)} & 91.78 & 0.1843 \\
\hline
\textbf{VIB (Beta=0.001)} & 91.49 & 0.1132 \\
\hline
\textbf{VIB (Beta=0.01)} & 86.42 & 0.0456 \\ 
\hline
\textbf{VIB (Beta=0.1)} & 78.31 & 0.0321 \\
\hline
\textbf{VIB (Beta=1)} & 9.67 & 0.0092 \\
\hline
\textbf{IBAL} & 91.32 & 0.0876  \\
\hline
\textbf{CLAD (Ours)} & \textbf{92.33} & \textbf{0.0471} \\ 
\hline
\end{tabular}
\label{CIFAR_Res}
\end{table}

\begin{figure}[ht]
    \centering
    \subfloat[Model inversion and attribute inference attacks at varying SNR.]{
        \includegraphics[width=0.85\linewidth]{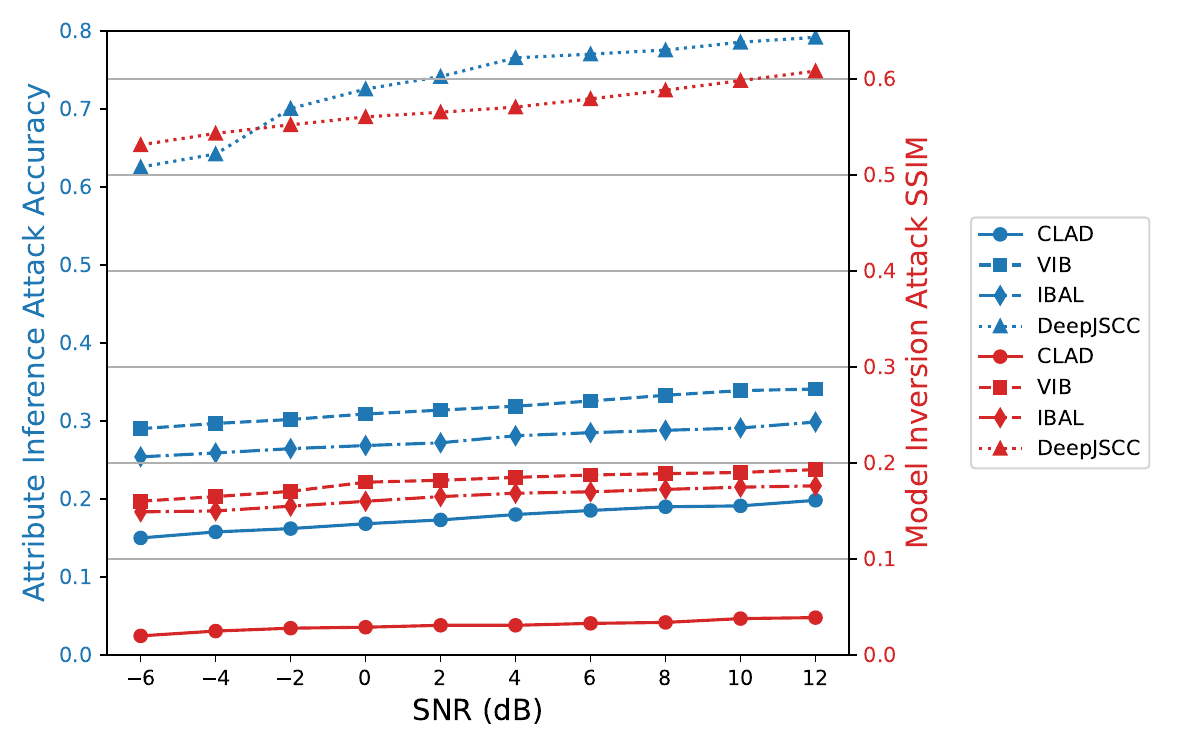}
        \label{fig:privacy_snr}
    }
    
    \vspace{0.8em}  

    \subfloat[Model inversion and attribute inference attacks under varying latencies with SNR = 12dB.]{
        \includegraphics[width=0.85\linewidth]{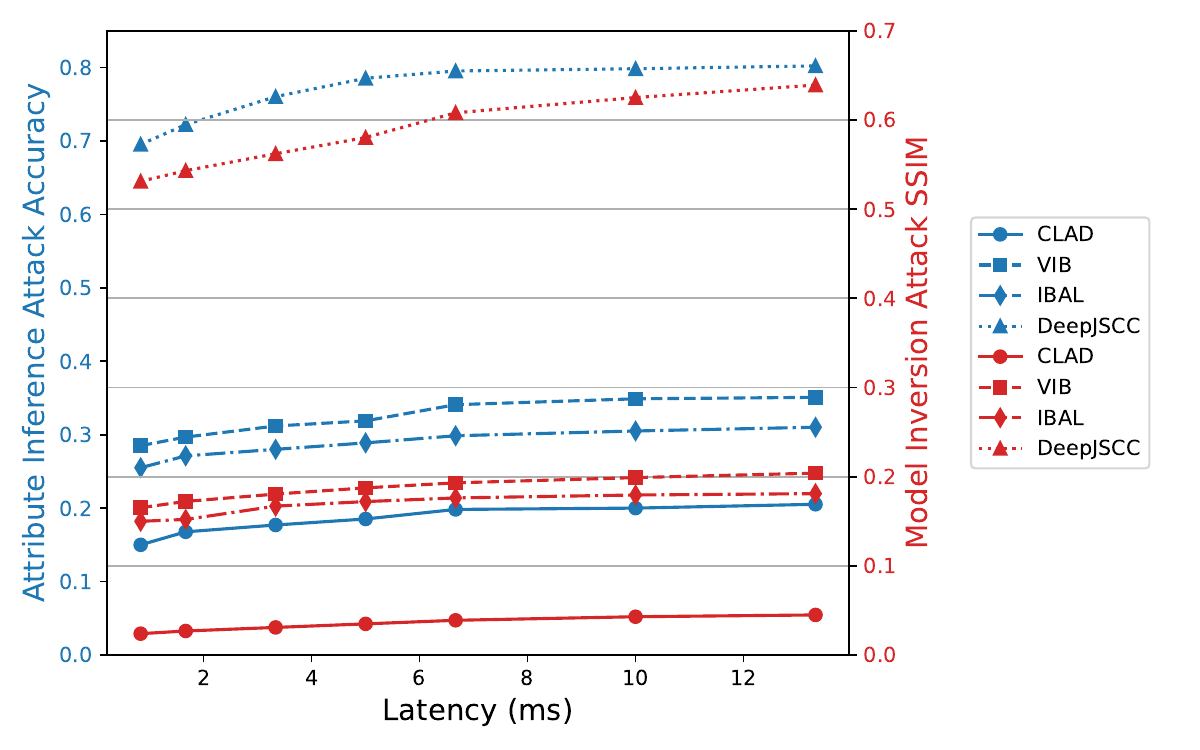}
        \label{fig:privacy_latency}
    }
    
    \caption{Privacy evaluation comparing CLAD, VIB, IBAL, and DeepJSCC across different settings on the Colored MNIST dataset.}
    \label{fig:privacy_stacked}
\end{figure}

\begin{figure}[!t]
    \centering
    \subfloat[t-SNE embedding for DeepJSCC\label{fig:tsne_deepjscc}]{
        \includegraphics[width=0.47\textwidth]{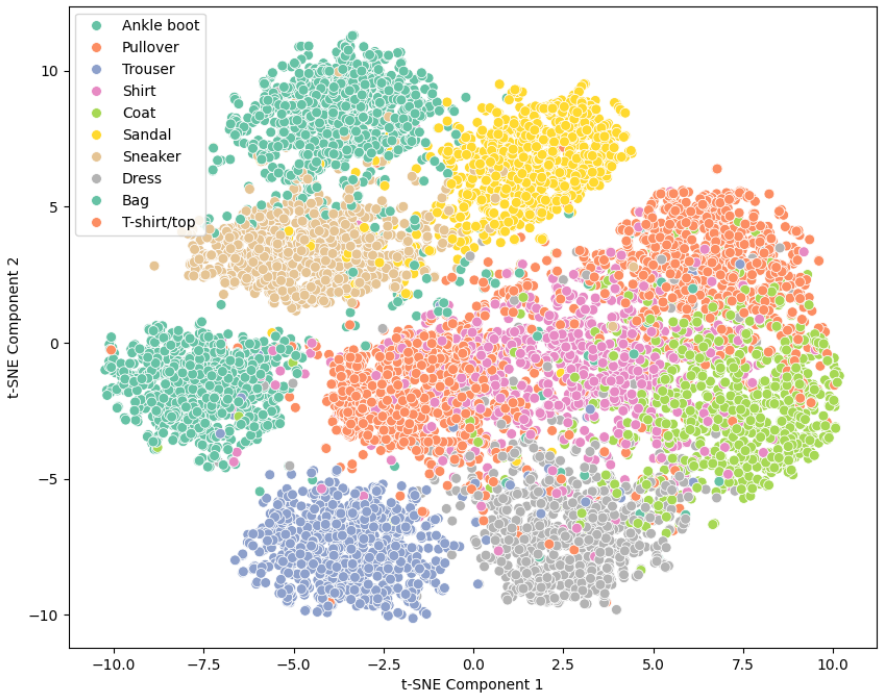}
    }\hfill
    \subfloat[t-SNE embedding for CLAD\label{fig:tsne_clad}]{
        \includegraphics[width=0.47\textwidth]{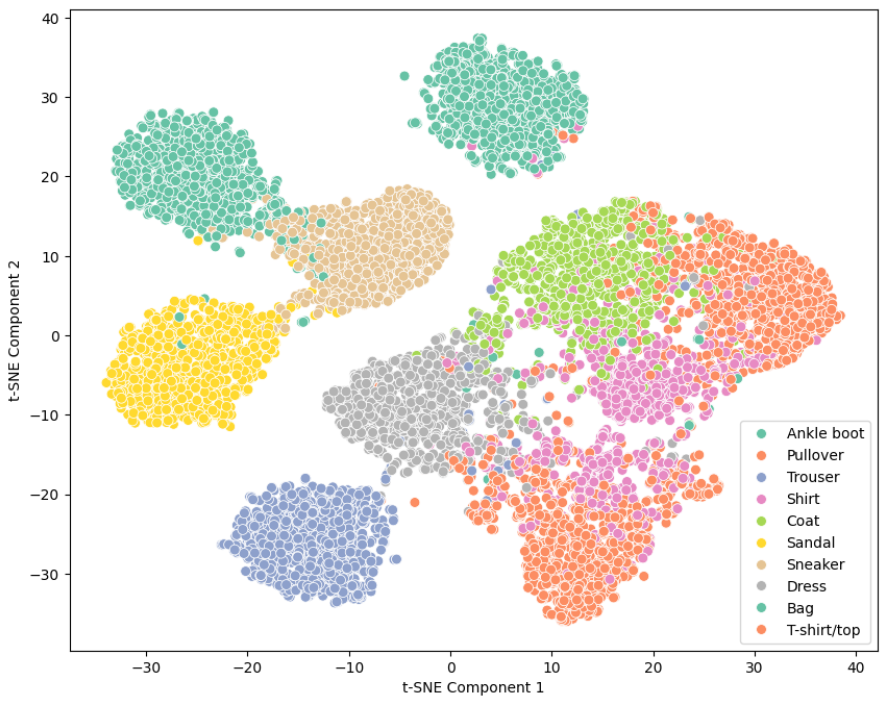}
    }
    \caption{2-dimensional t-SNE embeddings of the received feature representations for the Colored FashionMNIST classification task at SNR = 12 dB.}
    \label{fig:tsne}
\end{figure}

\begin{figure}[!t]
    \centering
    \subfloat[Colored MNIST Reconstructions\label{fig:colored_mnist_reconstructions}]{
        \includegraphics[width=0.47\textwidth]{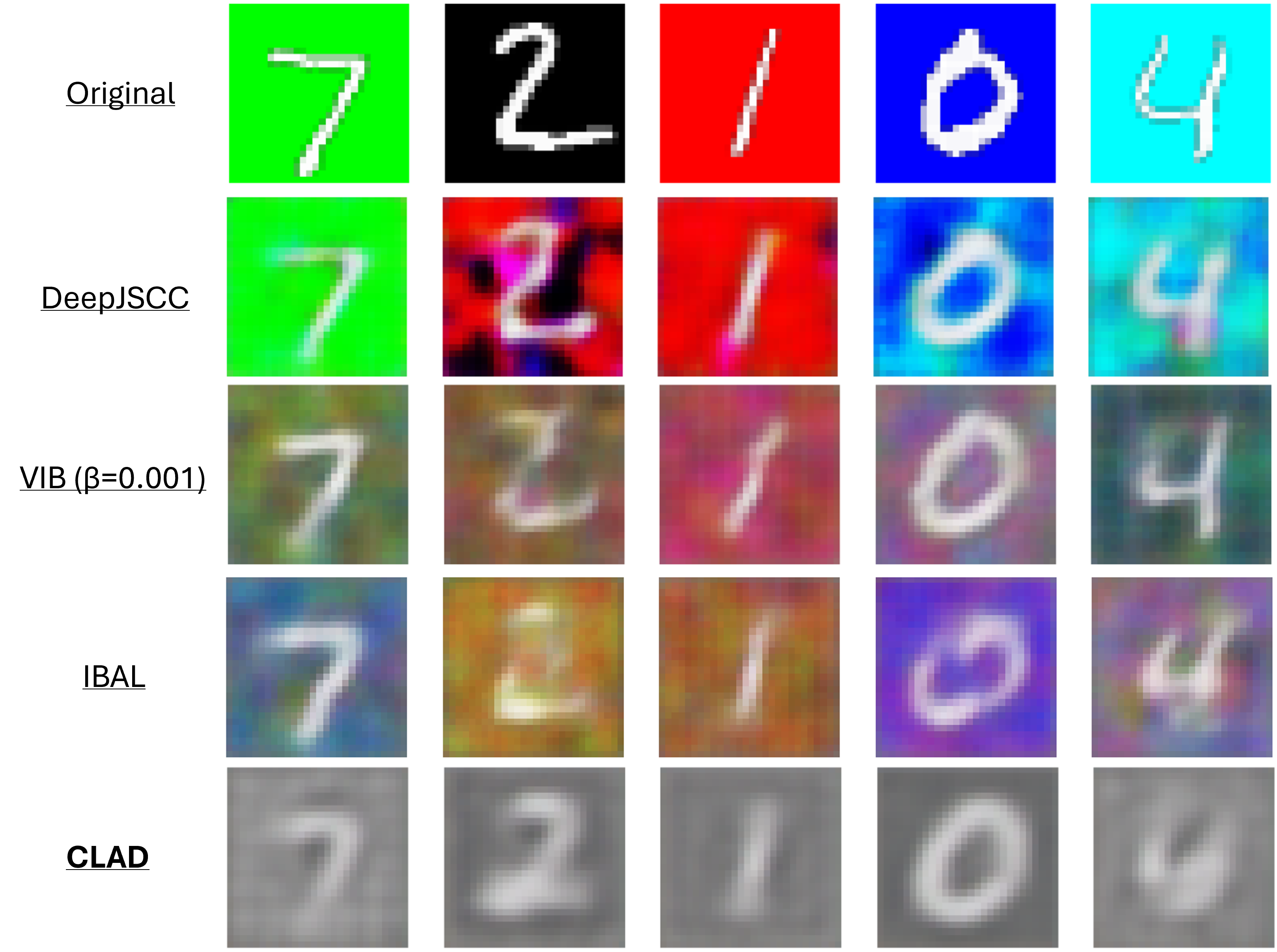}
    }\hfill
    \subfloat[Colored Fashion MNIST Reconstructions\label{fig:fmnist_reconstructions}]{
        \includegraphics[width=0.47\textwidth]{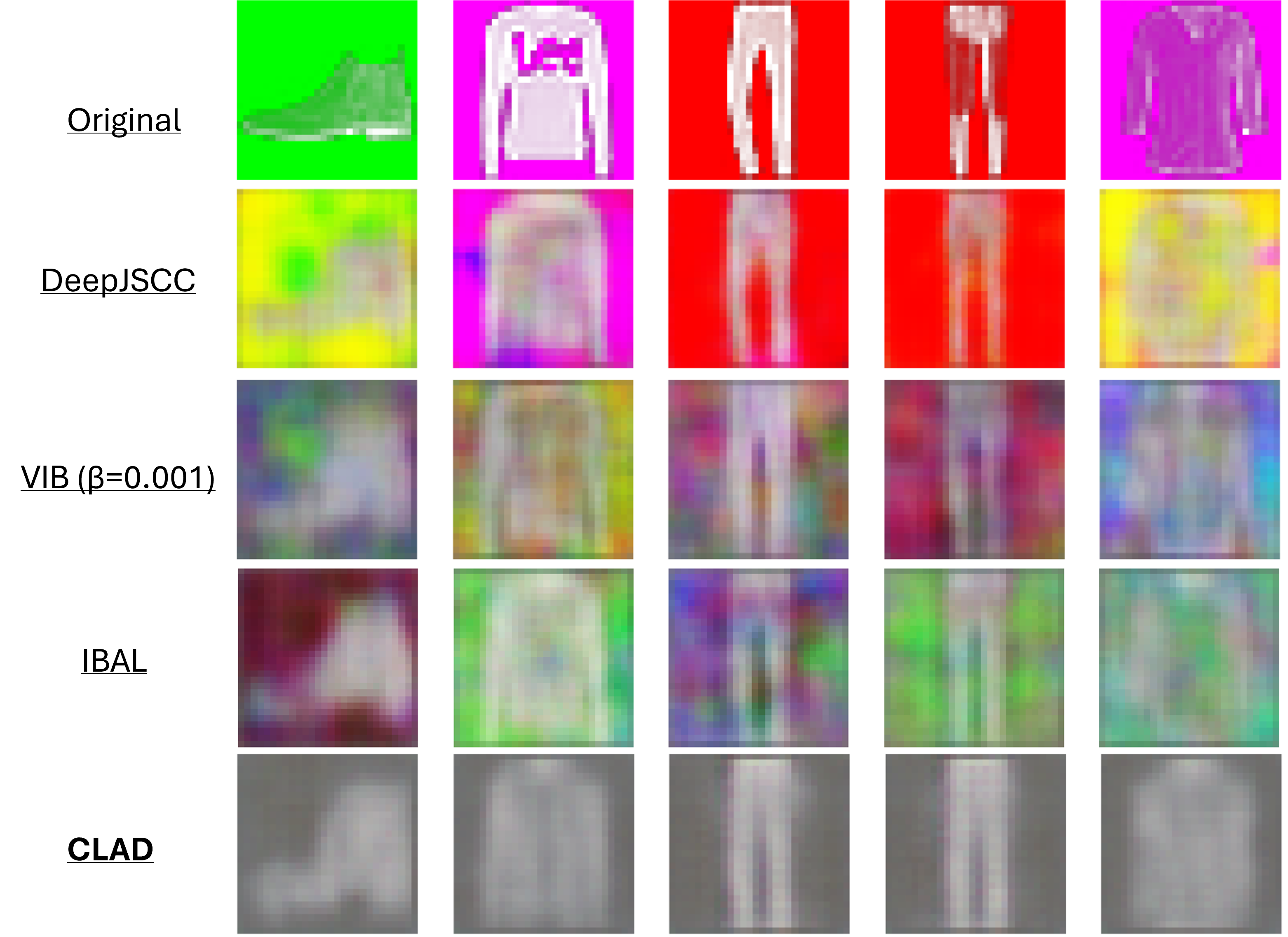}
    }
    \caption{Reconstructed images from the received feature representations under different methods.}
    \label{fig:reconstructions}
\end{figure}

\subsection{Results and Analysis}

In this subsection we thoroughly analyze and discuss the performance of the proposed method CLAD against the three baselines, DeepJSCC, VIB, and IBAL on all three aforementioned datasets.

\subsubsection{Task-Oriented Classification Performance}

We begin by evaluating the classification accuracy of each method at a fixed SNR of 12 dB under a latency constraint of $t \leq 6.67$ ms. The latency constraint can also be seen as an encoded feature vector with a maximum of 64 dimension. Tables~\ref{MNIST_RES}, \ref{FMNIST_Res}, and \ref{CIFAR_Res} summarize the results for Colored MNIST, Colored FashionMNIST, and CIFAR-10, respectively. CLAD consistently achieves the highest accuracy across all datasets, outperforming DeepJSCC, VIB (across various $\beta$ values), and IBAL. 

For instance, on Colored MNIST, CLAD achieves 98.42\% accuracy, surpassing DeepJSCC (97.96\%) and the best-performing VIB configuration (98.01\%) with better privacy metrics. On Colred FashionMNIST, CLAD outperforms all other baselines by 1.50-3.00\%. On the more complex CIFAR-10 dataset, CLAD achieves 92.33\%, outperforming DeepJSCC by 1.12\% and VIB ($\beta = 0.001$) by 0.84\%.

We further evaluate classification robustness under dynamic transmission scenarios. As shown in Fig.~\ref{fig:graph_results_1}, CLAD maintains higher accuracy across all SNR levels from -6 dB to 12 dB. This robustness is particularly crucial in 6G-IoT environments, where channel conditions fluctuate and low-latency, on-device inference is essential.

\subsubsection{Rate-Distortion Tradeoff under Latency Constraints}

In real-time 6G-IoT systems, minimizing latency while maintaining high task performance is critical. Since communication latency in our setup is determined by the dimension of the transmitted feature vector \( \hat{Z} \), which is fixed across methods, we assess how efficiently each method encodes task-relevant information within that constraint.

Fig.~\ref{fig:graph_results_2} shows the rate-distortion curves, where distortion corresponds to classification error and rate is reflected by the latency budget. Despite identical feature vector sizes, CLAD consistently achieves lower distortion. This indicates that CLAD produces more compact and informative semantic representations by focusing on preserving task-relevant features, as a result, CLAD achieves superior downstream accuracy under the same latency constraints, yielding a more favorable rate-distortion tradeoff. This makes it particularly well-suited for efficient, privacy-aware communication in latency-sensitive 6G-IoT applications.

\subsubsection{Privacy Evaluation: Information Retention, Attribute Inference, and Model Inversion}

Privacy is critical for trustworthy communication in 6G-IoT systems, particularly when transmitting semantically rich representations over shared or noisy channels. To assess this, we evaluate how well each method limits task-irrelevant information leakage under varying conditions and attacks.

\vspace{1mm}
\textbf{IRI and Attribute Inference.} From Tables~\ref{MNIST_RES} , \ref{FMNIST_Res} and \ref{CIFAR_Res}, CLAD achieves the lowest IRI and attribute inference accuracy across all methods and datasets. On Colored MNIST, CLAD reaches an IRI of 0.039 and adversarial accuracy of 19.83\%, significantly outperforming DeepJSCC (IRI of 0.608, adversarial accuracy 79.16\%) and VIB (ranging from 0.3931 to 0.0123 in IRI as $\beta$ increases). While higher $\beta$ values in VIB reduce information retention and adversarial success, they also degrade task accuracy dropping to 93.29\% at $\beta = 0.1$ and 11.36\% at $\beta = 1$. In contrast, CLAD maintains both privacy and classification accuracy (98.42\%). IBAL provides a lower IRI (0.1762) and adversarial accuracy compared to the best performing VIB results, however CLAD outperforms it in both privacy and task performance.

Similarly, on Colored FashionMNIST, CLAD achieves an IRI of 0.0587 and attribute inference accuracy of 19.33\%, outperforming VIB (0.2562 IRI and 47.00\% inference accuracy at $\beta=0.001$) and DeepJSCC (0.5958 IRI, 79.52\% attribute inference accuracy). IBAL shows a decent privacy tradeoff (IRI 0.1762, inference 29.86\% on MNIST), but CLAD surpasses it in both privacy and task performance. On the more complex dataset, CIFAR10, a similar trend is seen, as CLAD provides the best task accuracy and lowest IRI compared to all other baselines.

\vspace{1mm}
\textbf{Privacy vs. SNR and Latency.} The impact of varying SNR and latency on privacy is illustrated in Fig.~\ref{fig:privacy_stacked}. In Fig.~\ref{fig:privacy_snr}, attribute inference success increases with SNR for all methods, as better channel conditions enhance signal fidelity, but CLAD consistently maintains the lowest inference accuracy and model inversion SSIM. For instance, at 12 dB, DeepJSCC shows attribute inference accuracy above 75\% and SSIM exceeding 0.6, while CLAD remains below attribute inference accuracy 25\% and under 0.05 SSIM. IBAL and VIB offer intermediate privacy performance, but are sitll outperformed by CLAD. Fig.~\ref{fig:privacy_latency} presents a comparison of privacy across different latency levels at a fixed SNR of 12 dB. Across all latency settings, CLAD consistently achieves the lowest attribute inference accuracy and SSIM, indicating strong resistance to both attacks.

\subsubsection{Visual Analysis and Qualitative Comparisons}

In addition to numerical metrics, Fig.~\ref{fig:reconstructions} presents reconstructed images under each method. CLAD effectively removes stylistic and task-irrelevant details (e.g., color, background texture), focusing on shape and structure relevant for classification. VIB and IBAL reconstructions are blurrier, while DeepJSCC retains vivid background and texture details, making it vulnerable to inference attacks. These visual insights further support the quantitative findings. Furthermore, Fig.~\ref{fig:tsne} shows a 2D t-SNE visualization of the encoded features for the Colored FashionMNIST dataset. The embeddings produced by CLAD result in more compact and clearly separated class clusters compared to DeepJSCC, indicating better task relevance and intra-class consistency. In contrast, DeepJSCC features show more overlap and spread.

\section{Conclusions}

We proposed CLAD, a task-oriented communication framework that combines contrastive learning and adversarial disentanglement to extract compact, task-relevant features while improving privacy. By reducing task-irrelevant information in the latent space, CLAD enhances downstream performance and strengthens resistance to attribute inference and model inversion attacks. We also introduced the IRI, a practical and reproducible metric for comparing information preservation across methods. Extensive experiments across multiple datasets and channel conditions demonstrate that CLAD outperforms state-of-the-art baselines in terms of task accuracy, privacy, and minimality of representation, making it a suitable solution for semantic communication in 6G-IoT systems. Future work may explore extending CLAD to more complex input modalities and dynamically adaptive scenarios where task relevance can shift over time.

\bibliographystyle{IEEEtran}

\bibliography{references}





\end{document}